\pdfoutput=1

\documentclass[11pt]{article}

\usepackage[final]{acl}

\usepackage{times}
\usepackage{latexsym}

\usepackage[T1]{fontenc}

\usepackage[utf8]{inputenc}

\usepackage{microtype}

\usepackage{inconsolata}

\usepackage{graphicx}

\usepackage[normalem]{ulem}
\usepackage{enumitem}
\usepackage{multirow}
\usepackage{booktabs}

\newcommand{\classifier}{\textsc{CAspER}}

\usepackage[most]{tcolorbox}
\newtcolorbox{mybox}[1][]{%
  colback=green!5!white,    
  colframe=green!40!black,  
  fonttitle=\bfseries,
  title=#1,
  boxrule=0.8pt,           
  arc=2mm,                 
  left=2mm,
  right=2mm,
  top=1mm,
  bottom=1mm
}

\usepackage{todonotes}

\title{\classifier\ in the Machine: Insights into Character Variety in LLM-Generated Stories}

\author{
  Anneliese Brei$^1$ \quad 
  Abhisheik Sharma$^2$  \quad
  \textbf{Nicholas Sanaie$^1$} \quad \\  
  \textbf{Lu Wang$^3$} \quad
  \textbf{Snigdha Chaturvedi$^1$}  \\
  $^1$UNC Chapel Hill  \quad 
  $^2$Georgia Institute of Technology \quad 
  $^3$University of Michigan\\
  \texttt{\href{mailto:abrei@cs.unc.edu}{abrei@cs.unc.edu}},\quad
  \texttt{\href{mailto:asharma914@gatech.edu}{asharma914@gatech.edu}},\quad
  \texttt{\href{mailto:nsanaie@unc.edu}{nsanaie@unc.edu}},\\
  \texttt{\href{mailto:wangluxy@umich.edu}{wangluxy@umich.edu}},\quad
  \texttt{\href{mailto:snigdha@cs.unc.edu}{snigdha@cs.unc.edu}}
  }

\begin{document}
\maketitle
\begin{abstract}
As LLM-generated text is increasingly used, especially in fictional domains, we explore how much LLM-generated stories differ from human-written stories. In this work, we focus on characters. We borrow definitions from narratology to analyze 8 intricate category-pairs of character, such as \textit{stylization} and \textit{wholeness}. These category-pairs consider more than just basic characteristics. They assess how characters are portrayed within their stories. After automatically inferring categories of characters within both LLM and human-written stories, we compare and contrast these two sets of stories. We consider the following overarching questions: (1) Do LLMs and human-written stories have similar characters? and (2) Do LLMs generate stories with a variety of characters? Our analysis includes research questions that focus on stories generated by popular LLMs and recently published human-written stories. We describe a number of interesting similarities, differences and key takeaways.\footnote{All code and annotated data is released publicly: \url{https://github.com/adbrei/casper}}
\end{abstract}

\section{Introduction}

Increasing numbers of authors are using AI tools to assist in the process of writing stories \cite{mirowski2024robot,chakrabarty2024creativity,mirowski2023co,ippolito2022creative,yuan2022wordcraft}. For example, Large Language Models (LLMs) are highly versatile for personalization, making their use ideal for tailoring writing to specific needs \cite{chakrabarty2024can,wasi2024ink,nicolicioiu2024panza}. 
In the digital humanities, LLMs play an important role for creative writing by assisting with textual analysis, interpretation, and generation \cite{cigliano2024impact}. 
While LLMs continue to improve at story generation, many writers and readers wonder whether or not LLMs are capable of generating interesting stories like humans. This multi-faceted question is complicated to address.

\begin{figure}[t]
    \centering
    \includegraphics[width=\linewidth]{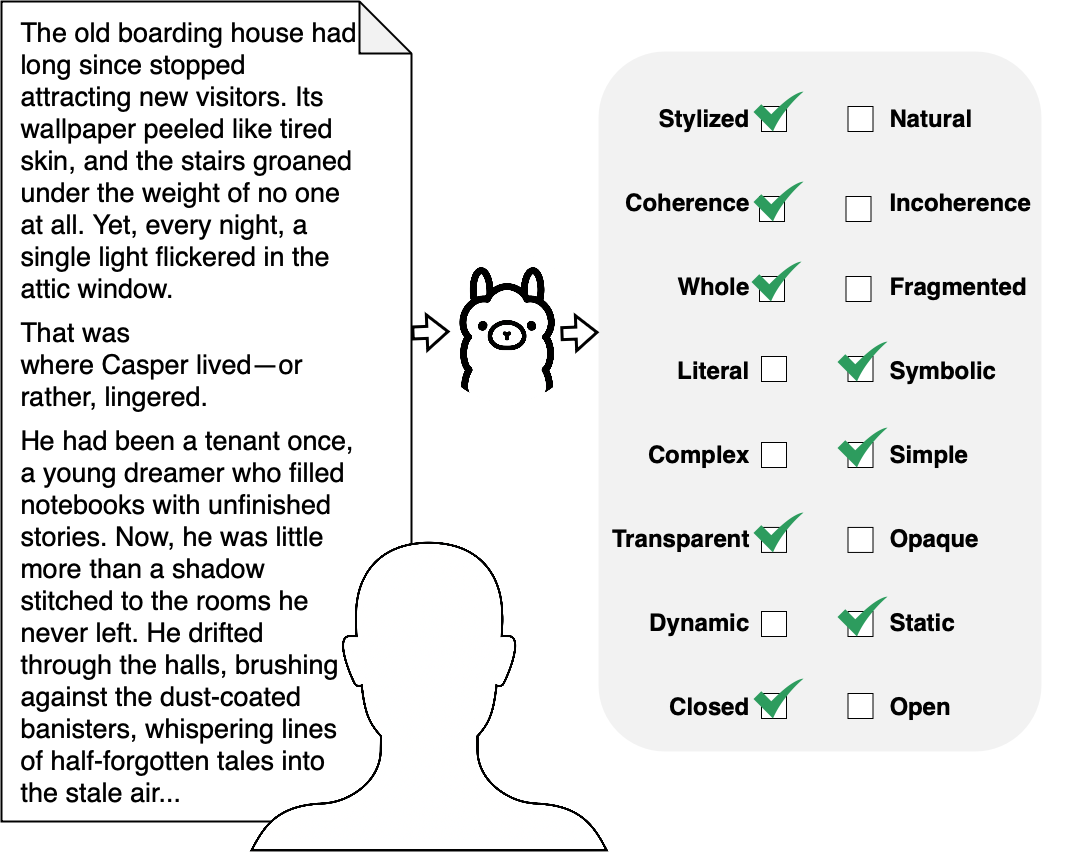}
    \caption{We analyze characters in LLM-generated short stories (left) using 8 category-pairs (right) that consider how characters are portrayed. 
    (e.g., a character might be represented in a \textit{fragmented} way, setting the tone for a disjointed mood within the story). 
    We compare LLM-generated characters with human-written characters other and LLM-generated characters from different model sizes and families.}
    \label{fig:task_intro}
\end{figure}

Narratologists have established that character and plot are two integral aspects of story \cite{janko1987aristotle,campbell2008hero}. In fact, theorists increasingly emphasize that the evaluation of character should be given additional priority because the role and function of character helps to define other elements of the story such as themes, tone, and culture \cite{bal1997narratology,forster1927aspects}. For example, if a narrative sequence is repeated among stories, but the characters themselves are different, then the overall takeaway messages of these stories are likely to differ \cite{phelan1989reading,propp1968morphology}. However, while recent work analyzes LLM-generated stories with respect to narrative development via plot \cite{tian2024large}, as of now, no existing works explore these stories' portrayal of character at a level deeper than personality traits.

In this paper, we investigate how LLMs represent character in stories they generate by asking, \textit{Do LLMs generate stories with a variety of characters?} This question is challenging because it cannot be answered by looking only at basic character attributes like demographics or basic personality. Instead, there must be a deeper understanding of how the character is portrayed.
To analyze characters within the light of their presentation style in a wholesome way, \citet{forster1927aspects} famously coins the terms ``flat'' versus ``round'' characters, where either type of character is ultimately determined by how the character is presented to the reader. However, though narratologists attempt to understand and define these terms, to this day the terms remain vague and subjective to interpretation \cite{jannidis2019character,phelan2005living,bal1997narratology,chatman1978story,booth1983rhetoric}. As such, the definitions of ``flat'' versus ``round'' are not easy to operationalization automatically.

With this in mind, we turn to a related character classification \cite{hochman1985character} inspired by \citet{forster1927aspects}'s analysis of character. This classification is a taxonomy that looks at finer and more automatically measurable categories in a flexible and more concrete fashion \cite{fishelov1990types}. 
It considers eight primary categories of character portrayal, including \textit{stylization},
\textit{coherency}, \textit{wholeness}, \textit{literalness}, \textit{complexity}, \textit{transparency}, \textit{dynamism}, and \textit{closure}. We design a framework, \textit{\underline{C}h\underline{a}racter'\underline{s} \underline{P}ortrayal Classifi\underline{er}} (\classifier), 
that considers these categories.
By comparing each primary category with its opposite (e.g., \textit{stylization} vs. \textit{naturalism}), we create a metric to evaluate characters according to factors of storytelling.
For example, a character's \textit{transparency} relies on how their thoughts and motives are presented to the reader \cite{fishelov1990types}.
Thus, we are enabled to perform more nuanced explorations of the types of characters in generated stories, as shown in Figure \ref{fig:task_intro}.

Using these category-pairs, we compare characters from human-written and LLM-generated stories and analyze the distributions of character category-pairs from a variety of popular LLM families and sizes. We explore 6 research questions that address (1) the similarities and differences of LLM and human-written stories, using both coarse and fine-grained levels of comparison and (2) how LLM-generated stories differ, such as across model size and family, genre of story, and across multiple inference calls.
Answering these questions helps us to determine if LLM-generated characters have patterns with respect to other LLM or human-written characters, as well as the diversity of LLM generations. We provide key takeways from these research questions in Section \ref{sec:takeaways}. Our primary contributions include:
\begin{itemize}[topsep=1pt, leftmargin=*, noitemsep] 
    \item We adapt and modify theory-grounded methods of character analysis to automatically understand characters in LLM-generated stories. Our framework can be used to keep track of how future LLM-generated stories portray character;
    \item We construct a high-quality dataset of human-written and LLM-generated short stories that underwent careful curation to ensure comparability across genres and themes. It will be publicly released for future research. 
    \item We compare characters in generated stories to characters in human-written and other generated stories and provide a detailed analysis of similarities and differences.
\end{itemize}

\begin{table*}[t]
\centering
\footnotesize
\begin{tabular}{p{0.005\textwidth} p{0.45\textwidth} p{0.45\textwidth}}
\toprule
&\textbf{Primary Category} & \textbf{Opposing Category}\\
\midrule
1&\textbf{Stylization}: A character is depicted through deliberate idealization, idealization, exaggeration, or conventional artistic patterns, often emphasizing form or artifice.
&\textbf{Naturalism}: A character is depicted with close attention to ordinary human traits and behaviors, aiming for lifelike accuracy without obvious artistic distortion.\\
\midrule
2&\textbf{Coherence}: A character’s actions, speech, and inner life align in ways that form a consistent and logically understandable pattern.
&\textbf{Incoherence}: A character’s actions, speech, or inner life lack consistency, producing contradictions, unpredictability, or fragmentation.\\
\midrule
3&\textbf{Wholeness}: A character is presented as a fully integrated being, with sufficient detail provided to give the sense of a complete personality.
&\textbf{Fragmentariness}: A character is presented only in partial aspects, leaving significant gaps in their personality, background, or presence.\\
\midrule
4&\textbf{Literalness}: A character functions solely as an individual within the story world, without additional layers of meaning attached.
&\textbf{Symbolism}: A character functions as both an individual and as a representation of an abstract idea, theme, or cultural concept beyond the story.\\
\midrule
5&\textbf{Complexity}: A character shows multiple, sometimes conflicting traits, desires, or motivations that interact in nuanced ways.
&\textbf{Simplicity}: A character is defined by one or very few dominant traits or motivations, with little internal tension.\\
\midrule
6&\textbf{Transparency}: A character’s inner life—thoughts, emotions, and motivations—is made clear to the reader, often through narration or explicit cues.
&\textbf{Opacity}: A character’s inner life remains hidden, ambiguous, or difficult to interpret, leaving the reader uncertain about their drives or reasoning.\\
\midrule
7&\textbf{Dynamism}: A character undergoes noticeable development or transformation over the course of the narrative.
&\textbf{Staticism}: A character remains fundamentally unchanged in outlook, behavior, or values throughout the narrative.\\
\midrule
8&\textbf{Closure}: A character’s narrative arc is brought to a clear resolution, with all questions about them settled by the story’s end.
&\textbf{Openness}: A character’s narrative arc remains unresolved, with key aspects of their fate, choices, or significance left indeterminate.\\
\bottomrule
\end{tabular}
\caption{Eight category-pairs with definitions. Primary-category (left) is contrasted by opposing-category (right).}
\label{tab:category_pairs_defs}
\end{table*}

\section{Related Works}

Narrative theory is used in natural language processing to provide a theoretical framework for computational narrative understanding \cite{piper2021narrative}. As LLMs show increasingly impressive capabilities for text generation, much work attempts to distinguish LLM from human-written text \cite{boutadjine2025human,russell2025people,elek2025evaluating,godghase2025distinguishing,ali2025hlu,najjar2025leveraging,venkatraman2024collabstory,harrag2020bert,uchendu2024catch,gehrmann2019gltr}. Some works perform deeper analyses of specific qualities of LLM-generated text, such as analyzing discourse similarities and linguistic features \cite{namuduri2025qudsim,reinhart2025llms,10295100} and narrative elements such as plot \cite{tian2024large}.

Prior work on automatically understanding characters within text reuse methods appropriate for studying real humans, such as personality profiling \cite{shu2024llm,jiang2024personallm} and tracking emotions \cite{brahman2020modeling,rahimtoroghi2017modelling,chaturvedi2016ask}. Other work analyzes gender bias \cite{lucy2021gender,huang2021uncovering}, roles \cite{jang2024evaluating,bamman2013learning}, relationships between characters \cite{vijjini2022towards,kim2019frowning,chaturvedi2017unsupervised,iyyer2016feuding,srivastava2016inferring} character setting \cite{soni2023grounding}, and traits that relate characters to humans in society \cite{yuan2024evaluating,jaipersaud2024show,brahman-etal-2021-characters-tell,Yu_2024_ToM,li-etal-2023-multi-level,yu-etal-2023-personality,bamman2019annotated}.

However, these works do not consider how a fictional character is portrayed within the artistic framework of the story, which is the focus of this work. We refer to narratological discussions to explore how character development and story shape the reader's perception of the character \cite{smith2022engaging,carter2014characterization, janko1987aristotle, propp1968morphology}. In particular, we focus on categorization of character, which best known from \citet{forster1927aspects} and further developed by later narratologists \cite{phelan2005living, rimmon2003narrative, bal1997narratology, fishelov1990types, hochman1985character, chatman1978story, booth1983rhetoric}.

\section{\classifier\ Framework}
\label{sec:categorizing-aspects}
In this section we define categories, formulate the task, and find the best framework.

\subsection{Defining Category-Pairs}
\label{sec:defining_aspects_of_fictional_characters}

In order to evaluate character as a component of story-telling within story, we consider 8 category-pairs, described in Table \ref{tab:category_pairs_defs} \cite{hochman1985character}. We choose this framework instead of others that define ``flat'' versus ``round'' characters,\footnote{For example, we consider \citet{forster1927aspects}'s predominant definitions of ``flat'' (``built around a single idea or quality; they are predictable and do not change'') and ``round'' (``complex, multi-dimensional, and capable of surprising the reader''). Such definitions are coarse and often tricky to distinguish.}\footnote{Though \citet{hochman1985character} does not explicitly tie the framework to definitions of "round" vs. "flat" characters, other narratologists state that both classifications categorize characters for the same objective, with \citet{hochman1985character}'s taxonomy showing desirable flexiblility \cite{jannidis2019character,fishelov1990types}.} because these categories have more tangible definitions and are better suited for classification with LLMs.

Each category-pair consists of a category compared to its opposite. For example, character \textit{stylization} is contrasted with \textit{naturalism}, \textit{coherence} with \textit{incoherence}, \textit{wholeness} with \textit{fragmentariness}, and so forth. Human writers may choose the extreme into which their characters fall; hence in the real-world we find characters fitting every category. 

To further illustrate these category-pairs, we provide examples from popular fiction. 
Stereotypically, fairy tales and moral-centric stories contain \textit{stylized} characters \cite{jung1968archetypes,pearson1991awakening}, such as "the Innocent" (often a naive young girl with ample trust and virtue who eventually learns maturity) or "the Sage" (often an old man with a long beard who currently has profound knowledge, yet seeks deeper knowledge). In the Harry Potter series \cite{rowling1997harrypotterseries}, Albus Dumbledore fits the \textit{stylized} archetype of the wise sage with profound knowledge. Similarly, the Dursleys demonstrate \textit{stylization} because they represent mediocrity through exaggerated domestic satire.
On the other hand, Molly Weasley is \textit{natural} because she is portrayed as a mother with authentic motivations, actions and feelings (e.g., humor, love, pride, frustration, fear, anger). Hermione Granger is \textit{coherent} because her motives are revealed, and her actions make sense in light of her personality. 
Further examples for the rest of the category-pairs are given in Appendix \ref{sec:popular-fiction-examples}, and examples from our corpus are given in Appendix \ref{sec:examples_of_char}.

With each of these category-pairs, we propose \textit{\underline{C}h\underline{a}racter'\underline{s} \underline{P}ortrayal Classifi\underline{er}} (\classifier) for automatically understanding characters in stories. 

\subsection{Task Formulation}

As presented in Figure \ref{fig:task_intro}, we aim to classify characters using the category-pairs in Section \ref{sec:defining_aspects_of_fictional_characters}. Given a short story with character $c$, we consider the eight pairs: $\mathcal{A}_c=\langle A_{Sty.},$ $A_{Coh.},$ $A_{Whol.},$ $A_{Lit.},$ $A_{Comp.},$ $A_{Tran.},$ $A_{Dyn.},$ $A_{Clos.}\rangle$. Each component $A$ is evaluated by determining if $c$ best fits the category (e.g., $A_{Sty.}$ indicates $c$ is \textit{stylized}) or its opposite (e.g., $\bar A_{Sty.}$ indicates $c$ is \textit{natural}).

\begin{figure*}[h]
    \centering
    \includegraphics[width=\linewidth,keepaspectratio]{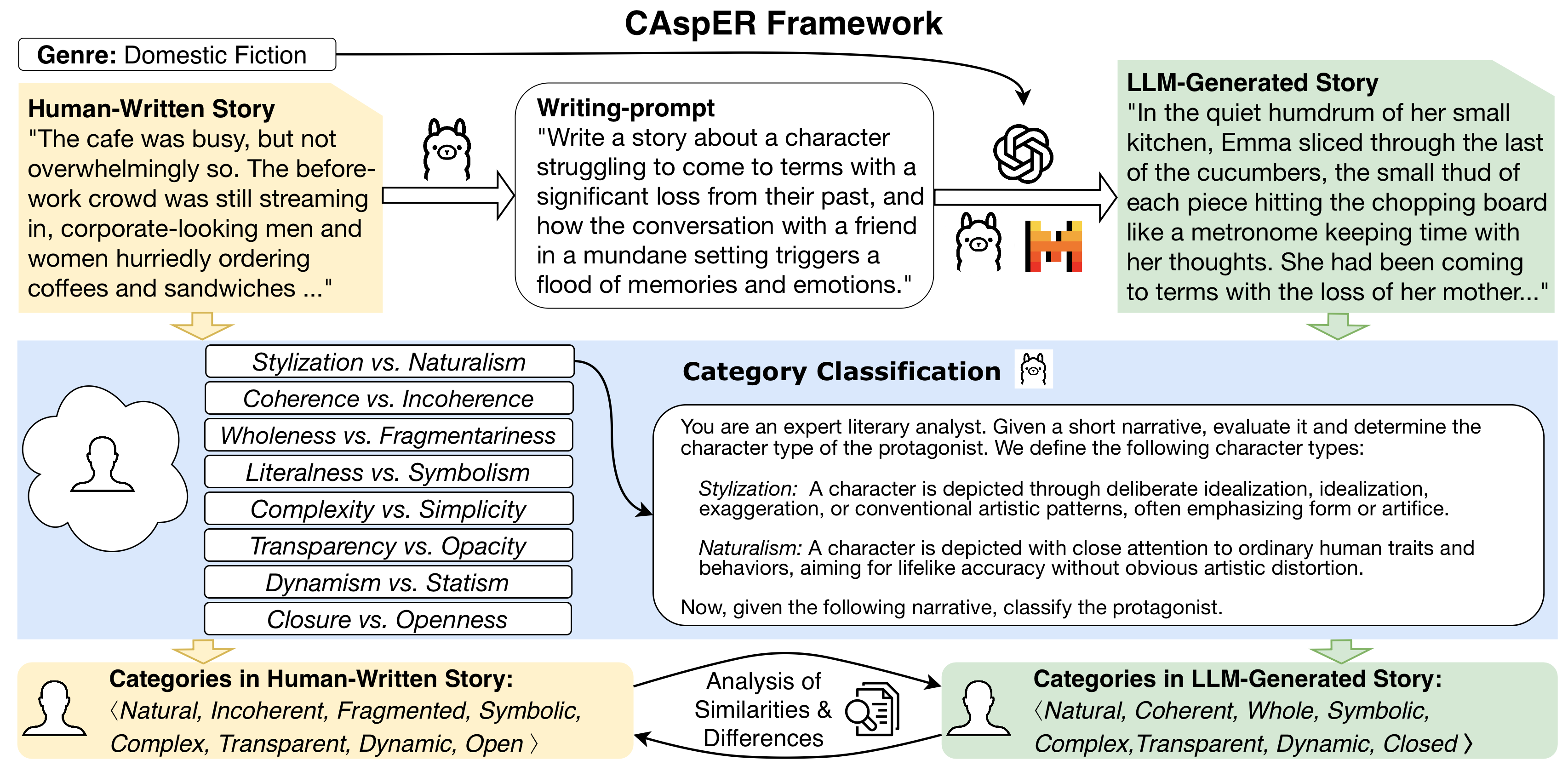}
    \caption{Overview of \classifier\ framework, including the pipeline for the creation of the corpus (top row), experiments (middle row), and analysis of the categories (bottom row).}
    \label{fig:pipeline}
\end{figure*}

\subsection{Identifying Categories of Character} 
\label{sec:evaluation-framework} 
 
First, we describe how we classify characters. Manual classification is costly and time-consuming due to necessary literary expertise and high text volume. To perform the intended analysis at scale, we need an automatic classification method. We choose to follow recent works using LLM-as-a-judge \cite{zheng2023judging} which allows us to do classifications at scale and inexpensively. 
However, this method has its own challenges because the task can be subjective and definitions can be difficult to interpret. Additionally, judges sometimes struggle with forms of bias and lack of knowledge within a specialized context \cite{li2025generation}. Since
\classifier\ uses categories specific to narratology, we observe a judge could have difficulty understanding literary nuances. 
To mitigate these potential issues, we test numerous settings using multiple models, different forms of definitions, various prompts and zero-shot and in-context learning (ICL) \cite{dong2024survey}. We pick the best setup by comparing the results to human annotations.

For task formulation, we (a) classify category-pairs individually, using one template for each pair; (b) classify all category-pairs in a single inference call; (c)  try a Likert-scale rating to evaluate category-pairs in a more fine-grained manner.

For describing category labels, we use (a) definitions that paraphrase the explanations in \cite{hochman1985character}, 
(b) descriptive adjective lists\footnote{The adjective list is a list of synonyms, effectively summarizing each category in a manner akin to a thesaurus lookup. 
E.g., \textit{transparent}: clear, obvious, apparent, straightforward, readable, comprehensible, explicit, direct, and understandable.
}, 
or (c) the combination of definitions and adjective lists.

For settings with ICL, we experiment with several template layouts, including (a) a ``basic'' layout that describes the categories before providing an example of each; (b) an ``interleaved'' layout that provides an example of a category immediately following the respective description; and (c) a ``repeated'' layout that first describes the categories, provides examples, and then repeats the same description of the categories. 
The formatting of our templates are given in Appendix \ref{append:evaluating_prompting_methods}.

To test these templates, we create a test set of LLM and human-written short stories.
We collect 50 human-written stories and 50 corresponding writing-prompts from \texttt{r/WritingPrompts}\footnote{https://www.reddit.com/r/WritingPrompts/}. From the writing-prompts, we generate 50 new stories with \texttt{GPT4o-mini}. All 100 stories are annotated for each category-pair by experts who are native English speakers and are familiar with the research and literary theory. The resulting Cohen's kappa inter-rater agreement for all category-pairs is $\kappa=0.56$, showing moderate agreement \cite{landis1977measurement} (more details in Appendix \ref{sec:inter-rater-agreement}).
We keep the size of our task manageable by considering only protagonists of stories since protagonists drive story flow and largely define narrative shape and genre. For these reasons, we consider only protagonists for the other experiments as well.

Finally, we evaluate our templates on the labeled stories in our test set using our largest open-sourced LLM, \texttt{Llama-70B}. Note, we also experiment with \texttt{GPT4o-mini} and find that both LLMs give comparable results (see Appendix \ref{sub:results-experiments-templates}).
We compare Macro F1-scores between the gold labels and predicted labels, and we determine that it is best to use a binary classification setup with separate inference calls for each category using only the definitions of the category-pairs. While ICL-interleaved performed well for all category pairs, to maximize performance, we choose the best-performing method for each category-pair: \textbf{ICL-repeated} for \textit{stylization}, \textbf{ICL-interleaved} for \textit{coherence}, \textit{wholeness}, \textit{complexity}, and \textit{transparency}, \textbf{ICL-basic} for \textit{literalness}, and \textbf{Zero-shot-basic} for \textit{dynamism} and \textit{closure}. A description of the results and F1 scores breakdown is given in Appendix \ref{sub:results-experiments-templates}.

\section{Evaluating Stories}

In this section, we describe our process of obtaining short stories generated by humans and LLMs for \classifier.
Then, using the best templates described in Section \ref{sec:evaluation-framework},  we perform experiments to understand the types of characters in our corpus.

\subsection{Corpus Creation}

For a broad representation of stories that might be found in the wild and to ensure comparability of stories, we seek stories that fit into the following 4 widely encompassing genres \cite{fong2013you}: (1) \textit{Domestic}: explores everyday life, family, relationships, and personal conflicts; (2) \textit{Romance}: focuses on a romantic relationship; (3) \textit{Science-Fiction/Fantasy}: emphasizes futuristic technology or magical realms; (4) \textit{Suspense/Thriller}: maintains a sense of urgency and tension through plots by putting characters into imminent danger.

We collect short stories written by humans from \texttt{r/shortstories}\footnote{ https://www.reddit.com/r/shortstories/}, one of the largest and most popular subreddits for writers. We take multiple steps to ensure that this collection is high-quality. First, we only consider submissions from the past year to avoid stories potentially used as training data for LLMs. Second, we conduct experiments to ensure these stories are (1) from the same textual domain as the stories in our test set from Section \ref{sec:evaluation-framework} and (2) not likely to be LLM-generated (see Appendix \ref{sec:comparining_human_written_stories}). Third, all stories have been previously tagged with genre labels by their authors (see Appendix \ref{append:mapping_of_genre_labels} for the mapping of subreddit genre labels to our genre definitions). To ensure these genres are not ambiguous, we use \texttt{Llama-70B} to classify the stories and eliminate all whose predicted genres do not match the subreddit labels. In this way we collect a total of 200 high-quality human-written stories (50 stories per genre).

We obtain short stories generated by LLMs in a two-step process, and we take multiple steps to ensure that these stories are directly comparable to the human-written stories. First, in order to elicit LLM-generated stories that are of comparable theme to the human-written stories, we use our largest open-source model, \texttt{Llama-70B}, to generate writing-prompts from the human-written stories. To optimize comparability across genres, we also specify that the writing-prompt should be appropriate for the genre of the human-written story. We use the following template, and manually verify reasonable writing-prompts are produced:

\begin{mybox}
For this story, give the best writing-prompt from which this story is written (i.e., it should be obvious the story is written from this writing-prompt). Please ensure that it is particularly suitable for the genre, \{\textit{genre}\}.
\end{mybox} 

Finally, for comparability across technical settings, for all open-sourced LLMs we generate short stories from each writing-prompt 3 times:

\begin{mybox}
You are a creative\footnote{We test the word ``creative'' does not bias the LLM towards generating characters of a certain type (e.g., more dramatic characters). See Appendix \ref{sec:appendix_checking_creativity}.} story writer. Use the writing-prompt below to generate a complete short story in the following genre: \{\textit{genre}\}.
\end{mybox} 

Our final corpus contains 200 human-written and 4400 LLM-generated stories. See Table \ref{tab:corpus_statistics} in Appendix \ref{sec:addition-details-corpus} for corpus statistics.

\subsection{Setup and Experiments}

We perform a comprehensive analysis that explores a broad range of LLM-generated short stories using popular LLMs of 8 different sizes from 4 families. We test one closed-source (\texttt{GPT4o-mini}) and 7 open-source LLMs (\texttt{Llama-3B}, \texttt{Llama-8B}, \texttt{Llama-70B}, \texttt{Phi3-4B}, \texttt{Phi3-14B}, \texttt{Mistral-7B}, \texttt{Mistral-24B}).\footnote{Model names/details are given in Appendix \ref{sec:implementation-details}, Table \ref{tab:model_families}.} Since we seek creative outputs while maintaining consistent model settings for evaluation purposes, for all inference we set $temperature=0.7$ and $top$-$p=0.9$ for open-sourced LLMs and default hyperparameters for \texttt{GPT4o-mini}. To allow flexibility in story-length and encourage story completion, we set the number of possible output tokens to the max permitted. A subset of outputs are hand-evaluated to ensure high-quality classifications.

\section{Analysis of Character Portrayal}
\label{sec:analysis}
To understand characters using \classifier, we consider 6 research questions (RQs) that compare (1) LLM vs. human-written stories and (2) LLM vs. other LLM-generated stories. \\

\noindent\textbf{\textit{(RQ1): How does the average LLM-generated character compare to the average human-written character?}} To observe categories of an ``average'' story, we determine which categories are most represented within a set of stories. We classify labels and rewrite $\mathcal{A}=\langle A_{Sty.},$ $A_{Coh.},$ $A_{Whol.},$ $A_{Lit.},$ $A_{Comp.},$ $A_{Tran.},$ $A_{Dyn.},$ $A_{Clos.}\rangle$, such that each $A$ is represented by majority labels (a label is in the majority if it makes up for $\ge 50$\% of the set of stories). For human-written stories, we observe 
$\mathcal{A}=\langle \bar A_{Sty.},$ $A_{Coh.},$ $A_{Whol.},$ $A_{Lit.},$ $A_{Comp.},$ $A_{Tran.},$ $A_{Dyn.},$ $\bar A_{Clos.}\rangle$.
For LLM-generated stories (averaged across outputs from all models), we observe 
$\mathcal{A}=\langle \bar A_{Sty.},$ $A_{Coh.},$ $A_{Whol.},$ $A_{Lit.},$ $A_{Comp.},$ $A_{Tran.},$ $A_{Dyn.},$ $A_{Clos.}\rangle$.
These vectors differ only for \textit{Closure}. This finding indicates that, though the majority of the human-written stories present characters with unfilled aspects at the end of the story, LLMs are more likely to generate characters who have a definite conclusion. 
We conclude that human-writers are likely to take more artistic liberties than LLMs by using ambiguity. In this comparison, LLM's instead tend to ``play it safe'' and tie up loose ends.\\

\begin{figure}[t]
    \centering
    \includegraphics[width=\linewidth]{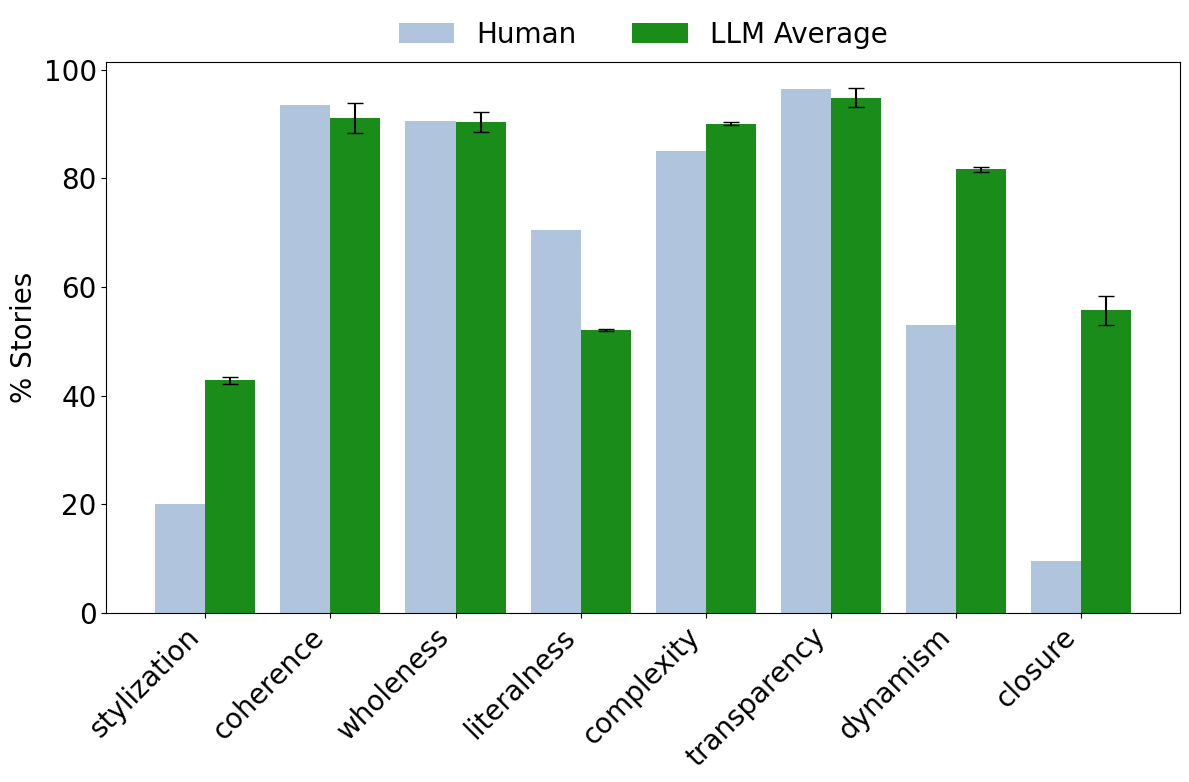}
    \caption{\textbf{RQ2:} We compare the percentage of human-written stories (blue) with the percentage of all LLM-generated stories (green) w.r.t. categories. For the latter, we average generations from all LLMs and show standard deviation across LLMs with vertical lines.}
    \label{fig:rq2}
\end{figure}

\noindent\textbf{\textit{(RQ2): How do LLM-generated characters compare to human-written characters in a more fine-grained manner?}} To better understand the remaining binary values from \textit{RQ1} in a more fine-grained analysis, we compare the percentage of categories in human vs. LLM-generated stories across all models. 
Figure \ref{fig:rq2} shows the results. We observe LLM-generated characters are statistically more likely to be \textit{stylized} and \textit{dynamic} than human-written characters. In other words, LLMs prefer to create stereotypical characters who grow over the course of a story. This pattern likely comes from the models’ tendency to write characters whose emotions shift noticeably as the narrative unfolds (e.g., a character who begins the story feeling sad but ends feeling joyful). Such characters who fit archetypes often help convey a moral lesson. This conjecture is supported by the observation that LLM's also have fewer \textit{literal} (more \textit{symbolic}) characters than human-written stories, pointing to the presence of underlying themes and takeaway lessons hidden within the story. Meanwhile, Figure \ref{fig:rq2} shows both LLMs and humans overwhelmingly prefer \textit{coherent}, \textit{whole}, and \textit{transparent} characters. In other words, characters are more likely to be straightforward and fully described to the reader. This finding is surprising because it indicates a lack in diversity of characters (i.e., there are fewer mysterious or confusing characters); however, it might be a result of focusing on protagonists, who are often the most developed characters. \\

\begin{figure}[t]
    \centering
    \includegraphics[width=\linewidth]{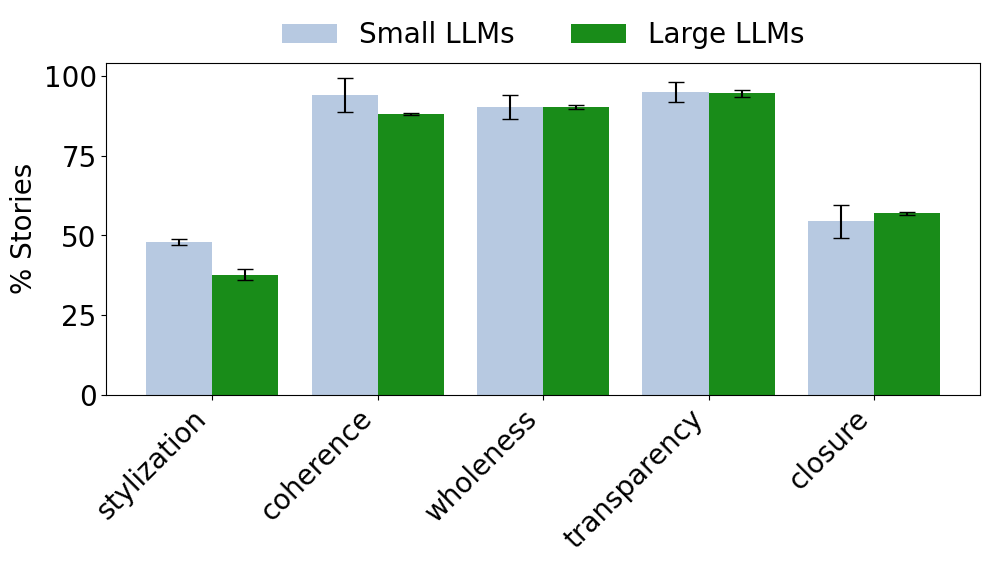}
    \caption{\textbf{RQ3}: To determine if primary-categories of character have trends according to model size only, we compare the average of small-medium LLMs (grey) and large LLMs (green) w.r.t. primary categories of character. Standard deviations shown as vertical bars.}
    \label{fig:rq3}
\end{figure}

\noindent\textbf{\textit{(RQ3) Does LLM size affect character portrayal?}}
Next, we focus on characters produced by smaller vs. larger LLMs. We ask, do larger LLMs ($\ge 14B$) generate characters with more varied distributions than smaller LLMs ($<14B$)? 
We compare \textit{stylization}, \textit{coherence}, \textit{wholeness}, \textit{transparency}, and \textit{closure} which show the most standard deviation in Figure ~\ref{fig:rq2}. If larger models create more nuanced characters, we would expect distinct distributions across these categories. However, as shown in Figure~\ref{fig:rq3}, averages of smaller and larger LLMs are nearly identical. The only minor difference is that smaller LLMs generate  $10.2\%$ more \textit{stylized} characters. Our findings are surprising because they indicate that, in general, though larger LLMs are better at text generation, they do not necessarily produce different types of characters than smaller LLMs. Overall, model size does not appear to affect the types of characters produced.\\

\begin{figure}[t]
    \centering
    \includegraphics[width=\linewidth]{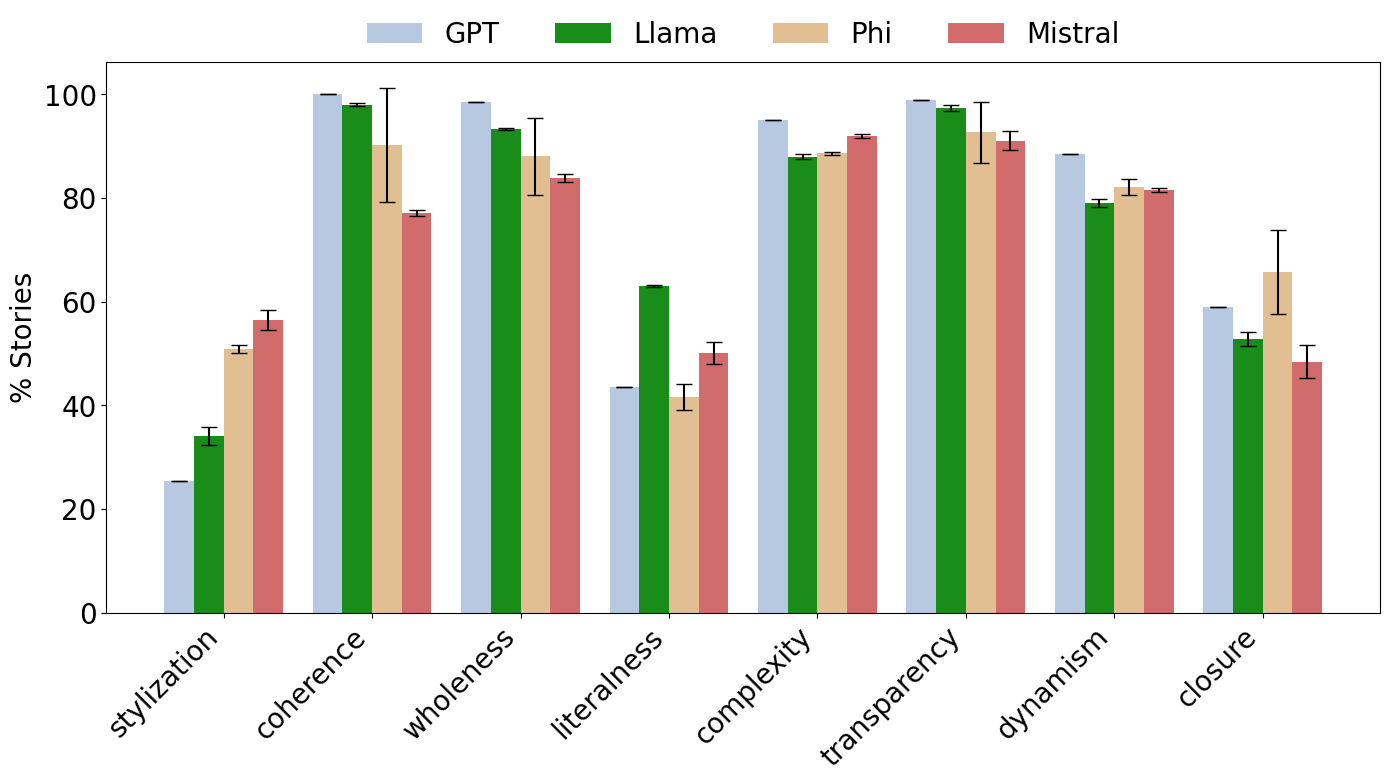}
    \caption{\textbf{RQ4:} Averaged categories of characters across families. Standard deviations across different-sized models within a family are shown as vertical bars.}
    \label{fig:rq4}
\end{figure}

\noindent\textbf{\textit{(RQ4): How does the diversity of categories vary among stories generated by different models?}}
We consider the distribution of aspects of character with respect to each family tested. Figure \ref{fig:rq4} shows the average distribution of category-pairs averaged for all LLMs belonging to a family. We observe average distributions follow general trends within each category. In other words, all families have \textit{whole} characters for $>80\%$ stories, \textit{literal} characters for $<64\%$ stories, etc. However, within these trends, we see the greatest amount of variations between families for \textit{stylization} and \textit{coherence}, \textit{literalness}, and \textit{closure}. Compared to the other families, Mistral prefers more \textit{stylized} and less \textit{coherent}, \textit{closed} characters, indicating an interesting mixture of exaggerated characters who might raise questions from readers. We notice only Phi shows significant standard deviation ($>5\%$) across models. Ignoring GPT (where only one model is evaluated), Llama shows the least amount of standard deviation. From these findings, we surmise that Phi generates the most diverse characters and Llama generates the least diverse characters.\\ 

\begin{figure}
    \centering
    \includegraphics[width=\linewidth]{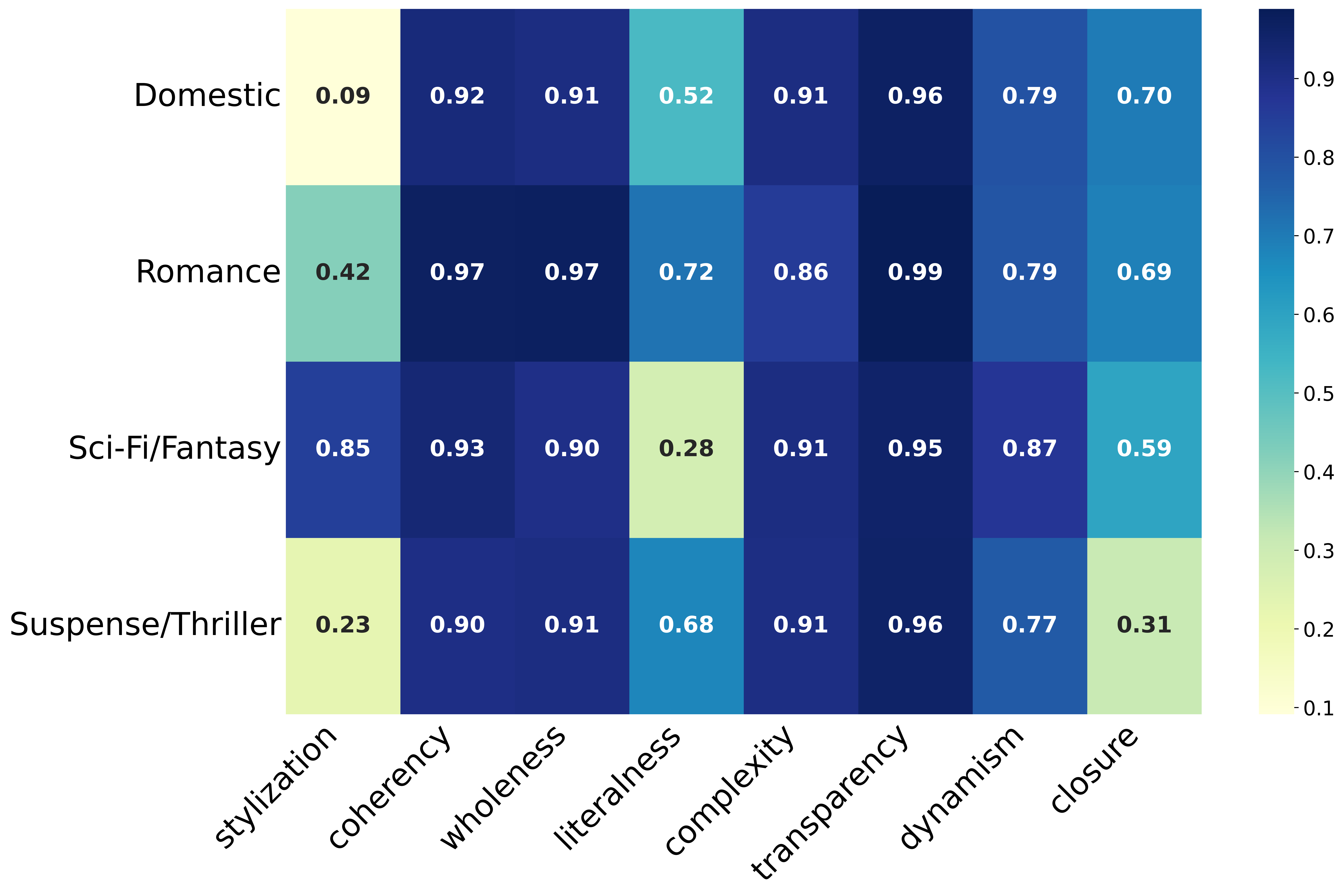}
    \caption{\textbf{RQ5:} Distribution of categories, given each specified genre across all LLM-generated stories.}
    \label{fig:rq5}
\end{figure}

\noindent\textbf{\textit{(RQ5) Do LLMs default to particular categories for different genres?}}
We analyze genres to see if they affect the type of character produced. Figure \ref{fig:rq5} shows the fraction of times a character fits a category, given the genre specified during story-generation.
Firstly, Domestic stories hardly have \textit{stylized} characters, which is suitable for life-like characters; however, these characters are \textit{literal} in only $52\%$ of stories. This is surprising because, even though the stories are focused on everyday life, these characters still represent abstract themes without relying on archetypes.
Romance stories have nearly completely \textit{coherent}, \textit{whole}, \textit{transparent} characters, which indicates the characters are straight-forward and not confused with emotions, as we might expect. 
Sci-Fi/Fantasy stories have the highest percentage of \textit{stylized} characters and symbolic characters, which makes sense, because this genre of stories is known to exaggerate conditions or be set in futuristic settings. Only $59\%$ of these characters are \textit{closed}, indicating that at the end of almost half these stories, there are unresolved questions about the character, meaning that these stories might encourage more reader's curiosity. 
Finally, Suspense/Thriller stories tend to have more \textit{natural} characters, possibly introducing the reader to a convincing character, thereby increasing tension to suspenseful situations. Almost all of these characters are \textit{coherent} and \textit{transparent}, which is surprising because suspense and thrill are often built upon mystery and subterfuge. However, in thrillers, if suspense does not come from the characters, it must come from other aspects of the story, such as setting or plot. Overall, we observe genre affects types of characters generated by LLMs. A corresponding heatmap for human-written stories is given in Appendix \ref{sec:genres-for-human-stories}.\\

\noindent\textbf{\textit{(RQ6) If we provide the same prompt multiple times to the same model, do categories vary meaningfully among generations?}}
We consider how categories vary across multiple story generations using the same writing-prompt for \textit{stylization}, \textit{literalness}, and \textit{closure} 
(graphs and details for all categories are given in Appendix \ref{sec:rq6_appendix}). We observe for each fore-mentioned category respectively 30\%, 13.5\% and 3\% of the writing-prompts yield characters with the same label across all inference calls.
Most noticeably -- compared to \textit{stylization} which has a large number of ``stable'' characters that do not change across inference calls -- \textit{literalness} and \textit{closure} demonstrate the most amount of variability compared to the rest of the categories. This is surprising because it shows LLMs pick up fewer context clues from the writing-prompts that dictate how \textit{literal} and \textit{closed} a character should be.

\section{Takeaways}
\label{sec:takeaways}
From Section \ref{sec:analysis}, we observe a few key takeaways:

\begin{itemize}[topsep=1pt, leftmargin=*, noitemsep]
    \item Unlike human-written stories, LLMs-generated stories are more likely to ``play it safe'' and have characters with completed story-lines.

    \item LLM-generated characters are more likely to show character-growth during the course of the story compared to human-written characters.


    \item Larger models do not necessarily generate different types of characters than smaller models.

    \item The Phi family appears to generate the most diverse characters, and the Llama family appears to generate the least diverse characters.

    \item Genre affects types of characters generated by LLMs (e.g., Domestic stories tend to have realistic characters that represent themes, and Romance stories have straightforward characters).
    
    \item When re-generating stories from one prompt, \textit{literalness} and \textit{closure} have most variability.

\end{itemize}

\section{Conclusion}
We propose \classifier\ for classifying character portrayal using 8 category-pairs to compare characters in LLM and human-written stories. We create a corpus of 4400 LLM-generated stories across 7 models from 4 model families. This corpus was carefully curated so that each story can be compared to a parallel human-written story with a common writing-prompt. We answer research questions that compare LLM-generated characters with other LLM and human-written characters, and we provide a list of key takeaways to better understand types of characters produced by LLMs. 

We note \classifier\ is designed with definitions from narrative theory for the classification of stories. We do not extend these definitions to a different textual domain because this change introduces complex new problems. In particular, we would have to consider whether or not the new textual domain contains enough aspects of narrative \cite{piper2023computational} to portray a character so that category-pairs such as \textit{stylization} and \textit{wholeness} can be analyzed. Furthermore, some categories might exhibit subtle differences in a non-creative story setting (e.g., a \textit{stylized} character might be different in a report than in a story). We propose the exploration of these ideas for future work.

\section*{Limitations}

Since it is difficult to find open-source human-written stories, we scrape stories from subreddits. However, it is possible that a small number of these ``human-written'' stories might be enhanced or written by AI. We attempt to mitigate such cases by using software for detecting machine-generated text, including \texttt{Binoculars}\ \cite{hans2024spotting} and Copyleaks AI Detector.\footnote{https://copyleaks.com/ai-content-detector} All stories are written in English and invite future work to analyze character aspects of generated stories in other languages.

Since we utilize writing-prompts from the creative writing domain to generate short stories, we note our research is limited to the domain of such writing-prompts and stories. 

Also, though we have done due diligence to improve classification performance as much as possible, we note that automatic classification remains worse than human annotators. Since the final corpus is large, we expect these errors to be smoothed out for the analysis. However, the classifier will remain imperfect.

\section*{Acknowledgments}
We are grateful to the anonymous reviewers.
This work was supported in part by NSF grant DRL2112635, IIS2047232, and the Air Force Office of Scientific Research grant FA9550-22-1-0099.

\bibliography{custom}

@inproceedings{li2025generation,
  title={From generation to judgment: Opportunities and challenges of llm-as-a-judge},
  author={Li, Dawei and Jiang, Bohan and Huang, Liangjie and Beigi, Alimohammad and Zhao, Chengshuai and Tan, Zhen and Bhattacharjee, Amrita and Jiang, Yuxuan and Chen, Canyu and Wu, Tianhao and others},
  booktitle={Proceedings of the 2025 Conference on Empirical Methods in Natural Language Processing},
  pages={2757--2791},
  year={2025}
}

@book{rowling1997harrypotterseries,
  author    = {J. K. Rowling},
  title     = {Harry Potter Series},
  year      = {1997--2007},
  publisher = {Bloomsbury},
  address   = {London},
  note      = {7 volumes published between 1997 and 2007}
}

@book{propp1968morphology,
  author    = {Vladimir Propp},
  title     = {Morphology of the Folktale},
  year      = {1968},
  publisher = {University of Texas Press},
  address   = {Austin},
  edition   = {2nd},
  note      = {Translated by Laurence Scott, edited by Louis A. Wagner, with an introduction by Alan Dundes}
}

@book{phelan1989reading,
  author    = {James Phelan},
  title     = {Reading People, Reading Plots: Character, Progression, and the Interpretation of Narrative},
  year      = {1989},
  publisher = {University of Chicago Press},
  address   = {Chicago}
}

@book{bal1997narratology,
  author    = {Mieke Bal},
  title     = {Narratology: Introduction to the Theory of Narrative},
  year      = {1997},
  edition   = {2nd},
  publisher = {University of Toronto Press},
  address   = {Toronto}
}

@book{forster1927aspects,
  title={Aspects of the Novel},
  author={Forster, Edward Morgan},
  year={1927},
  publisher={Harcourt, Brace}
}

@book{booth1983rhetoric,
  title={The rhetoric of fiction},
  author={Booth, Wayne C},
  year={1983},
  publisher={University of Chicago Press}
}

@book{chatman1978story,
  title={Story and discourse: Narrative structure in fiction and film},
  author={Chatman, Seymour Benjamin and Chatman, Seymour},
  year={1978},
  publisher={Cornell university press}
}

@book{rimmon2003narrative,
  title={Narrative fiction: Contemporary poetics},
  author={Rimmon-Kenan, Shlomith},
  year={2003},
  publisher={Routledge}
}

@book{phelan2005living,
  title={Living to tell about it: A rhetoric and ethics of character narration},
  author={Phelan, James},
  year={2005},
  publisher={Cornell University Press}
}

@article{fishelov1990types,
  title={Types of character, characteristics of types},
  author={Fishelov, David},
  journal={Style},
  pages={422--439},
  year={1990},
  publisher={JSTOR}
}

@incollection{jannidis2019character,
  author    = {Jannidis, Fotis},
  title     = {Character},
  booktitle = {the living handbook of narratology},
  editor    = {H{\"u}hn, Peter and others},
  publisher = {Hamburg University},
  address   = {Hamburg},
  year      = {2019},
  url       = {http://www.lhn.uni-hamburg.de/article/character},
  note      = {Viewed 12 Feb 2019}
}

@book{jung1968archetypes,
  author    = {C. G. Jung},
  editor    = {R. F. C. Hull},
  title     = {The Archetypes and the Collective Unconscious},
  edition   = {2nd Edition},
  year      = {1968},
  publisher = {Routledge},
  address   = {London},
  doi       = {10.4324/9781315725642},
  isbn      = {9781315725642},
  pages     = {560},
  note      = {First published 1968, eBook published 17 December 2014}
}

@article{pearson1991awakening,
  title={Awakening the heroes within: Twelve archetypes to help us find ourselves and transform our world},
  author={Pearson, Carol},
  journal={(No Title)},
  year={1991}
}

@incollection{carter2014characterization,
  title={Characterization},
  author={Carter, Candice C and Pickett, Linda},
  booktitle={Youth Literature for Peace Education},
  pages={25--42},
  year={2014},
  publisher={Springer}
}

@book{smith2022engaging,
  title     = {Engaging Characters: Fiction, Emotion, and the Cinema},
  author    = {Smith, Murray},
  year      = {2022},
  publisher = {Oxford University Press},
  address   = {Oxford, UK}
}

@article{landis1977measurement,
  title={The measurement of observer agreement for categorical data},
  author={Landis, JR and
          Koch, GG},
  journal={Biometrics},
  volume={33},
  number={1},
  pages={159174},
  year={1977}
}

@inproceedings{dong2024survey,
  title={A Survey on In-context Learning},
  author={Dong, Qingxiu and Li, Lei and Dai, Damai and Zheng, Ce and Ma, Jingyuan and Li, Rui and Xia, Heming and Xu, Jingjing and Wu, Zhiyong and Chang, Baobao and others},
  booktitle={Proceedings of the 2024 Conference on Empirical Methods in Natural Language Processing},
  pages={1107--1128},
  year={2024}
}

@article{ippolito2022creative,
  title={Creative writing with an ai-powered writing assistant: Perspectives from professional writers},
  author={Ippolito, Daphne and Yuan, Ann and Coenen, Andy and Burnam, Sehmon},
  journal={arXiv preprint arXiv:2211.05030},
  year={2022}
}

@inproceedings{chakrabarty2024creativity,
  title={Creativity Support in the Age of Large Language Models: An Empirical Study Involving Professional Writers},
  author={Chakrabarty, Tuhin and Padmakumar, Vishakh and Brahman, Faeze and Muresan, Smaranda},
  booktitle={Proceedings of the 16th Conference on Creativity \& Cognition},
  pages={132--155},
  year={2024}
}

@inproceedings{mirowski2024robot,
  title={A robot walks into a bar: Can language models serve as creativity supporttools for comedy? an evaluation of llms’ humour alignment with comedians},
  author={Mirowski, Piotr and Love, Juliette and Mathewson, Kory and Mohamed, Shakir},
  booktitle={Proceedings of the 2024 ACM Conference on Fairness, Accountability, and Transparency},
  pages={1622--1636},
  year={2024}
}

@inproceedings{mirowski2023co,
  title={Co-writing screenplays and theatre scripts with language models: Evaluation by industry professionals},
  author={Mirowski, Piotr and Mathewson, Kory W and Pittman, Jaylen and Evans, Richard},
  booktitle={Proceedings of the 2023 CHI conference on human factors in computing systems},
  pages={1--34},
  year={2023}
}

@inproceedings{yuan2022wordcraft,
  title={Wordcraft: story writing with large language models},
  author={Yuan, Ann and Coenen, Andy and Reif, Emily and Ippolito, Daphne},
  booktitle={Proceedings of the 27th International Conference on Intelligent User Interfaces},
  pages={841--852},
  year={2022}
}

@article{chakrabarty2024can,
  title={Can AI writing be salvaged? Mitigating Idiosyncrasies and Improving Human-AI Alignment in the Writing Process through Edits},
  author={Chakrabarty, Tuhin and Laban, Philippe and Wu, Chien-Sheng},
  journal={arXiv preprint arXiv:2409.14509},
  year={2024}
}

@inproceedings{wasi2024ink,
  title={Ink and individuality: Crafting a personalised narrative in the age of llms},
  author={Wasi, Azmine Toushik and Islam, Raima and Islam, Mst Rafia},
  booktitle={Proceedings of the Third Workshop on Intelligent and Interactive Writing Assistants},
  pages={43--47},
  year={2024}
}

@article{nicolicioiu2024panza,
  title={Panza: Design and analysis of a fully-local personalized text writing assistant},
  author={Nicolicioiu, Armand and Iofinova, Eugenia and Jovanovic, Andrej and Kurtic, Eldar and Nikdan, Mahdi and Panferov, Andrei and Markov, Ilia and Shavit, Nir and Alistarh, Dan},
  journal={arXiv preprint arXiv:2407.10994},
  year={2024}
}

@book{janko1987aristotle,
  title={Aristotle: Poetics},
  author={Janko, Richard and others},
  year={1987},
  publisher={Hackett Publishing}
}

@book{campbell2008hero,
  title={The hero with a thousand faces},
  author={Campbell, Joseph},
  volume={17},
  year={2008},
  publisher={New World Library}
}

@book{hochman1985character,
  title        = {Character in Literature},
  author       = {Hochman, Baruch},
  year         = {1985},
  publisher    = {Cornell University Press},
  address      = {Ithaca, NY},
  isbn         = {0801417872, 9780801417870},
  pages        = {204},
}

@inproceedings{jang2024evaluating,
  title={Evaluating LLM Performance in Character Analysis: A Study of Artificial Beings in Recent Korean Science Fiction},
  author={Jang, Woori and Jung, Seohyon},
  booktitle={Proceedings of the 4th International Conference on Natural Language Processing for Digital Humanities},
  pages={339--351},
  year={2024}
}

@inproceedings{gehrmann2019gltr,
  title={GLTR: Statistical Detection and Visualization of Generated Text},
  author={Gehrmann, Sebastian and Strobelt, Hendrik and Rush, Alexander M},
  booktitle={Proceedings of the 57th Annual Meeting of the Association for Computational Linguistics: System Demonstrations},
  pages={111--116},
  year={2019}
}

@article{namuduri2025qudsim,
  title={QUDsim: Quantifying Discourse Similarities in LLM-Generated Text},
  author={Namuduri, Ramya and Wu, Yating and Zheng, Anshun Asher and Wadhwa, Manya and Durrett, Greg and Li, Junyi Jessy},
  journal={arXiv preprint arXiv:2504.09373},
  year={2025}
}

@inproceedings{lucy2021gender,
  title={Gender and representation bias in GPT-3 generated stories},
  author={Lucy, Li and Bamman, David},
  booktitle={Proceedings of the third workshop on narrative understanding},
  pages={48--55},
  year={2021}
}

@inproceedings{huang2021uncovering,
  title={Uncovering Implicit Gender Bias in Narratives through Commonsense Inference},
  author={Huang, Tenghao and Brahman, Faeze and Shwartz, Vered and Chaturvedi, Snigdha},
  booktitle={Findings of the Association for Computational Linguistics: EMNLP 2021},
  pages={3866--3873},
  year={2021}
}

@inproceedings{tian2024large,
  title={Are Large Language Models Capable of Generating Human-Level Narratives?},
  author={Tian, Yufei and Huang, Tenghao and Liu, Miri and Jiang, Derek and Spangher, Alexander and Chen, Muhao and May, Jonathan and Peng, Nanyun},
  booktitle={Proceedings of the 2024 Conference on Empirical Methods in Natural Language Processing},
  pages={17659--17681},
  year={2024}
}

@inproceedings{bamman2019annotated,
  title={An annotated dataset of literary entities},
  author={Bamman, David and Popat, Sejal and Shen, Sheng},
  booktitle={Proceedings of the 2019 Conference of the North American Chapter of the Association for Computational Linguistics: Human Language Technologies, Volume 1 (Long and Short Papers)},
  pages={2138--2144},
  year={2019}
}

@inproceedings{soni2023grounding,
  title={Grounding Characters and Places in Narrative Text},
  author={Soni, Sandeep and Sihra, Amanpreet and Evans, Elizabeth F and Wilkens, Matthew and Bamman, David},
  booktitle={The 61st Annual Meeting Of The Association For Computational Linguistics},
  year={2023}
}

@inproceedings{shu2024llm,
  title={When llm meets hypergraph: A sociological analysis on personality via online social networks},
  author={Shu, Zhiyao and Sun, Xiangguo and Cheng, Hong},
  booktitle={Proceedings of the 33rd ACM International Conference on Information and Knowledge Management},
  pages={2087--2096},
  year={2024}
}

@inproceedings{jiang2024personallm,
  title={PersonaLLM: Investigating the Ability of Large Language Models to Express Personality Traits},
  author={Jiang, Hang and Zhang, Xiajie and Cao, Xubo and Breazeal, Cynthia and Roy, Deb and Kabbara, Jad},
  booktitle={Findings of the Association for Computational Linguistics: NAACL 2024},
  pages={3605--3627},
  year={2024}
}

@inproceedings{yuan2024evaluating,
  title={Evaluating Character Understanding of Large Language Models via Character Profiling from Fictional Works},
  author={Yuan, Xinfeng and Yuan, Siyu and Cui, Yuhan and Lin, Tianhe and Wang, Xintao and Xu, Rui and Chen, Jiangjie and Yang, Deqing},
  booktitle={Proceedings of the 2024 Conference on Empirical Methods in Natural Language Processing},
  pages={8015--8036},
  year={2024}
}

@article{jaipersaud2024show,
  title={Show, Don't Tell: Uncovering Implicit Character Portrayal using LLMs},
  author={Jaipersaud, Brandon and Zhu, Zining and Rudzicz, Frank and Creager, Elliot},
  journal={arXiv preprint arXiv:2412.04576},
  year={2024}
}

@inproceedings{brahman-etal-2021-characters-tell,
    title = "{\textquotedblleft}Let Your Characters Tell Their Story{\textquotedblright}: A Dataset for Character-Centric Narrative Understanding",
    author = "Brahman, Faeze  and
      Huang, Meng  and
      Tafjord, Oyvind  and
      Zhao, Chao  and
      Sachan, Mrinmaya  and
      Chaturvedi, Snigdha",
    editor = "Moens, Marie-Francine  and
      Huang, Xuanjing  and
      Specia, Lucia  and
      Yih, Scott Wen-tau",
    booktitle = "Findings of the Association for Computational Linguistics: EMNLP 2021",
    month = nov,
    year = "2021",
    address = "Punta Cana, Dominican Republic",
    publisher = "Association for Computational Linguistics",
    url = "https://aclanthology.org/2021.findings-emnlp.150/",
    doi = "10.18653/v1/2021.findings-emnlp.150",
    pages = "1734--1752"
}

@inproceedings{Yu_2024_ToM,
author = {Yu, Mo and Wang, Qiujing and Zhang, Shunchi and Sang, Yisi and Pu, Kangsheng and Wei, Zekai and Wang, Han and Xu, Liyan and Li, Jing and Yu, Yue and Zhou, Jie},
title = {Few-shot character understanding in movies as an assessment to meta-learning of theory-of-mind},
year = {2024},
publisher = {JMLR.org},
booktitle = {Proceedings of the 41st International Conference on Machine Learning},
articleno = {2380},
numpages = {27},
location = {Vienna, Austria},
series = {ICML'24}
}

@inproceedings{li-etal-2023-multi-level,
    title = "Multi-level Contrastive Learning for Script-based Character Understanding",
    author = "Li, Dawei  and
      Zhang, Hengyuan  and
      Li, Yanran  and
      Yang, Shiping",
    editor = "Bouamor, Houda  and
      Pino, Juan  and
      Bali, Kalika",
    booktitle = "Proceedings of the 2023 Conference on Empirical Methods in Natural Language Processing",
    month = dec,
    year = "2023",
    address = "Singapore",
    publisher = "Association for Computational Linguistics",
    url = "https://aclanthology.org/2023.emnlp-main.366/",
    doi = "10.18653/v1/2023.emnlp-main.366",
    pages = "5995--6013"
}

@inproceedings{yu-etal-2023-personality,
    title = "Personality Understanding of Fictional Characters during Book Reading",
    author = "Yu, Mo  and
      Li, Jiangnan  and
      Yao, Shunyu  and
      Pang, Wenjie  and
      Zhou, Xiaochen  and
      Xiao, Zhou  and
      Meng, Fandong  and
      Zhou, Jie",
    editor = "Rogers, Anna  and
      Boyd-Graber, Jordan  and
      Okazaki, Naoaki",
    booktitle = "Proceedings of the 61st Annual Meeting of the Association for Computational Linguistics (Volume 1: Long Papers)",
    month = jul,
    year = "2023",
    address = "Toronto, Canada",
    publisher = "Association for Computational Linguistics",
    url = "https://aclanthology.org/2023.acl-long.826/",
    doi = "10.18653/v1/2023.acl-long.826",
    pages = "14784--14802"
}

@article{fong2013you,
  title={What you read matters: The role of fiction genre in predicting interpersonal sensitivity.},
  author={Fong, Katrina and Mullin, Justin B and Mar, Raymond A},
  journal={Psychology of aesthetics, creativity, and the arts},
  volume={7},
  number={4},
  pages={370},
  year={2013},
  publisher={Educational Publishing Foundation}
}

@inproceedings{cigliano2024impact,
  title={The Impact of Digital Analysis and Large Language Models in Digital Humanity},
  author={Cigliano, Andrea and Fallucchi, Francesca and Gerardi, Marco and others},
  booktitle={ICYRIME 2024: 9th International Conference of Yearly Reports on Infor-matics, Mathematics, and Engineering},
  pages={1},
  year={2024},
  organization={CEUR Workshop Proceedings}
}

@article{zheng2023judging,
  title={Judging llm-as-a-judge with mt-bench and chatbot arena},
  author={Zheng, Lianmin and Chiang, Wei-Lin and Sheng, Ying and Zhuang, Siyuan and Wu, Zhanghao and Zhuang, Yonghao and Lin, Zi and Li, Zhuohan and Li, Dacheng and Xing, Eric and others},
  journal={Advances in Neural Information Processing Systems},
  volume={36},
  pages={46595--46623},
  year={2023}
}

@misc{hans2024spotting,
          title={Spotting LLMs With Binoculars: Zero-Shot Detection of Machine-Generated Text}, 
          author={Abhimanyu Hans and Avi Schwarzschild and Valeriia Cherepanova and Hamid Kazemi and Aniruddha Saha and Micah Goldblum and Jonas Geiping and Tom Goldstein},
          year={2024},
          eprint={2401.12070},
          archivePrefix={arXiv},
          primaryClass={cs.CL}
    }

@inproceedings{piper2021narrative,
  title={Narrative theory for computational narrative understanding},
  author={Piper, Andrew and So, Richard Jean and Bamman, David},
  booktitle={Proceedings of the 2021 Conference on Empirical Methods in Natural Language Processing},
  pages={298--311},
  year={2021}
}

@inproceedings{piper2023computational,
  title={Computational narrative understanding: A big picture analysis},
  author={Piper, Andrew},
  booktitle={Proceedings of the Big Picture Workshop},
  pages={28--39},
  year={2023}
}

@article{reinhart2025llms,
  title={Do LLMs write like humans? Variation in grammatical and rhetorical styles},
  author={Reinhart, Alex and Markey, Ben and Laudenbach, Michael and Pantusen, Kachatad and Yurko, Ronald and Weinberg, Gordon and Brown, David West},
  journal={Proceedings of the National Academy of Sciences},
  volume={122},
  number={8},
  pages={e2422455122},
  year={2025},
  publisher={National Academy of Sciences}
}

@article{najjar2025leveraging,
  title={Leveraging Explainable AI for LLM Text Attribution: Differentiating Human-Written and Multiple LLMs-Generated Text},
  author={Najjar, Ayat and Ashqar, Huthaifa I and Darwish, Omar and Hammad, Eman},
  journal={arXiv preprint arXiv:2501.03212},
  year={2025}
}

@article{russell2025people,
  title={People who frequently use ChatGPT for writing tasks are accurate and robust detectors of AI-generated text},
  author={Russell, Jenna and Karpinska, Marzena and Iyyer, Mohit},
  journal={arXiv preprint arXiv:2501.15654},
  year={2025}
}

@article{boutadjine2025human,
  title={Human vs. machine: A comparative study on the detection of ai-generated content},
  author={Boutadjine, Amal and Harrag, Fouzi and Shaalan, Khaled},
  journal={ACM Transactions on Asian and Low-Resource Language Information Processing},
  volume={24},
  number={2},
  pages={1--26},
  year={2025},
  publisher={ACM New York, NY}
}

@inproceedings{ali2025hlu,
  title={HLU: Human Vs LLM Generated Text Detection Dataset for Urdu at Multiple Granularities},
  author={Ali, Iqra and Atuhurra, Jesse and Kamigaito, Hidetaka and Watanabe, Taro},
  booktitle={Proceedings of the 31st International Conference on Computational Linguistics},
  pages={3495--3510},
  year={2025}
}

@article{elek2025evaluating,
  title={Evaluating the Efficacy of Perplexity Scores in Distinguishing AI-Generated and Human-Written Abstracts},
  author={Elek, Alperen and Yildiz, Hatice Sude and Akca, Benan and Oren, Nisa Cem and Gundogdu, Batuhan},
  journal={Academic Radiology},
  year={2025},
  publisher={Elsevier}
}

@incollection{godghase2025distinguishing,
  title={Distinguishing chatbot from human},
  author={Godghase, Gauri Anil and Agrawal, Rishit and Obili, Tanush and Stamp, Mark},
  booktitle={Machine Learning, Deep Learning and AI for Cybersecurity},
  pages={529--564},
  year={2025},
  publisher={Springer}
}

@inproceedings{harrag2020bert,
  title={Bert Transformer model for Detecting Arabic GPT2 Auto-Generated Tweets},
  author={Harrag, Fouzi and Dabbah, Maria and Darwish, Kareem and Abdelali, Ahmed},
  booktitle={Proceedings of the Fifth Arabic Natural Language Processing Workshop},
  pages={207--214},
  year={2020}
}

@inproceedings{uchendu2024catch,
  title={Catch me if you gpt: Tutorial on deepfake texts},
  author={Uchendu, Adaku and Venkatraman, Saranya and Le, Thai and Lee, Dongwon},
  booktitle={Proceedings of the 2024 Conference of the North American Chapter of the Association for Computational Linguistics: Human Language Technologies (Volume 5: Tutorial Abstracts)},
  pages={1--7},
  year={2024}
}

@INPROCEEDINGS{10295100,
  author={Chong, Alicia Tsui Ying and Chua, Hui Na and Jasser, Muhammed Basheer and Wong, Richard T.K.},
  booktitle={2023 IEEE 13th International Conference on System Engineering and Technology (ICSET)}, 
  title={Bot or Human? Detection of DeepFake Text with Semantic, Emoji, Sentiment and Linguistic Features}, 
  year={2023},
  volume={},
  number={},
  pages={205-210},
  doi={10.1109/ICSET59111.2023.10295100}}

@article{venkatraman2024collabstory,
  title={Collabstory: Multi-llm collaborative story generation and authorship analysis},
  author={Venkatraman, Saranya and Tripto, Nafis Irtiza and Lee, Dongwon},
  journal={arXiv preprint arXiv:2406.12665},
  year={2024}
}

@inproceedings{brahman2020modeling,
  title={Modeling Protagonist Emotions for Emotion-Aware Storytelling},
  author={Brahman, Faeze and Chaturvedi, Snigdha},
  booktitle={Proceedings of the 2020 Conference on Empirical Methods in Natural Language Processing (EMNLP)},
  pages={5277--5294},
  year={2020}
}

@inproceedings{chaturvedi2016ask,
  title={Ask, and shall you receive? understanding desire fulfillment in natural language text},
  author={Chaturvedi, Snigdha and Goldwasser, Dan and Daume III, Hal},
  booktitle={Proceedings of the AAAI Conference on Artificial Intelligence},
  volume={30},
  number={1},
  year={2016}
}

@inproceedings{rahimtoroghi2017modelling,
  title={Modelling Protagonist Goals and Desires in First-Person Narrative},
  author={Rahimtoroghi, Elahe and Wu, Jiaqi and Wang, Ruimin and Anand, Pranav and Walker, Marilyn},
  booktitle={Proceedings of the 18th Annual SIGdial Meeting on Discourse and Dialogue},
  pages={360--369},
  year={2017}
}

@inproceedings{bamman2013learning,
  title={Learning latent personas of film characters},
  author={Bamman, David and O’Connor, Brendan and Smith, Noah A},
  booktitle={Proceedings of the 51st Annual Meeting of the Association for Computational Linguistics (Volume 1: Long Papers)},
  pages={352--361},
  year={2013}
}

@inproceedings{chaturvedi2017unsupervised,
  title={Unsupervised learning of evolving relationships between literary characters},
  author={Chaturvedi, Snigdha and Iyyer, Mohit and Daume III, Hal},
  booktitle={Proceedings of the AAAI Conference on Artificial Intelligence},
  volume={31},
  number={1},
  year={2017}
}

@inproceedings{iyyer2016feuding,
  title={Feuding families and former friends: Unsupervised learning for dynamic fictional relationships},
  author={Iyyer, Mohit and Guha, Anupam and Chaturvedi, Snigdha and Boyd-Graber, Jordan and Daum{\'e} III, Hal},
  booktitle={Proceedings of the 2016 Conference of the North American Chapter of the Association for Computational Linguistics: Human Language Technologies},
  pages={1534--1544},
  year={2016}
}

@inproceedings{srivastava2016inferring,
  title={Inferring interpersonal relations in narrative summaries},
  author={Srivastava, Shashank and Chaturvedi, Snigdha and Mitchell, Tom},
  booktitle={Proceedings of the AAAI Conference on Artificial Intelligence},
  volume={30},
  number={1},
  year={2016}
}

@inproceedings{vijjini2022towards,
  title={Towards Inter-character Relationship-driven Story Generation},
  author={Vijjini, Anvesh Rao and Brahman, Faeze and Chaturvedi, Snigdha},
  booktitle={EMNLP},
  year={2022}
}

@inproceedings{kim2019frowning,
  title={Frowning Frodo, Wincing Leia, and a Seriously Great Friendship: Learning to Classify Emotional Relationships of Fictional Characters},
  author={Kim, Evgeny and Klinger, Roman},
  booktitle={Proceedings of the 2019 Conference of the North American Chapter of the Association for Computational Linguistics: Human Language Technologies, Volume 1 (Long and Short Papers)},
  pages={647--653},
  year={2019}
}

@article{wolf2019huggingface,
  title={Huggingface's transformers: State-of-the-art natural language processing},
  author={Wolf, T},
  journal={arXiv preprint arXiv:1910.03771},
  year={2019}
}

@inproceedings{dror2018hitchhiker,
  title={The hitchhiker’s guide to testing statistical significance in natural language processing},
  author={Dror, Rotem and Baumer, Gili and Shlomov, Segev and Reichart, Roi},
  booktitle={Proceedings of the 56th annual meeting of the association for computational linguistics (volume 1: Long papers)},
  pages={1383--1392},
  year={2018}
}

\appendix

\section{Examples of Categories}
See Section \ref{sec:popular-fiction-examples} for popular fiction examples of characters who fit \classifier\ categories (Table \ref{tab:category_pairs_defs}). See Section \ref{sec:examples_of_char} for examples of stories from our corpus with protagonists fitting each category.
 
\subsection{Examples from Popular Fiction}
\label{sec:popular-fiction-examples}
In Section \ref{sec:defining_aspects_of_fictional_characters}, we give examples of \textit{stylized}, \textit{natural}, \textit{coherent}, \textit{incoherent}, \textit{whole}, and \textit{fragmented} characters from Harry Potter. Here, we continue the discussion and provide examples for the remaining category pairs:

Bellatrix Lestrange is \textit{incoherent} because her emotions swing unexpectedly between multiple forms of comic and tragic madness. 
Harry Potter is an example of a \textit{whole} character, whose entire background, motivations, ambitions, thoughts, and explanations for his actions are given to the reader. 
Sirius Black is \textit{fragmented} because his youth and years in Azkaban are not fully explained, though they influence his personality and decision making. 
Arthur Weasley is a \textit{literal} character with unique quirks and does not serve any known allegorical function beyond living daily life as a father. Dobby embodies the \textit{symbolic} character as an uncorrupted figure of innocence as he fights the oppression of servitude in search of freedom.
Professor Snape is one of the most \textit{complex} characters in the series because he balances extreme jealousy and love, thus sometimes showing cruelty though he is later proven to be caring. Rubeus Hagrid is a \textit{simple} character who only acts on loyalty and warm-heartedness towards his friends. Luna Lovegood is \textit{transparent} because her speech is honest without masking her inner thoughts. Albus Dumbledore is \textit{opaque} because he withholds information, and his motivations are layered and mostly hidden. Neville Longbottom is \textit{dynamic} because he starts as an awkward and anxious boy, and then he grows in courage to lead Dumbledore's Army and destroy the last Horcrux. Lord Voldemort is \textit{static} because his worldview remains constantly obsessed with power and death. Fred Weasley is \textit{closed}, not only because his death brings finality to his storyline, but also because there are no major unanswered questions about his role and life from the reader. Finally, Draco Malfoy is \textit{open} because he changes from childhood villain to disillusioned man with new but unclear morals and an undefined future.

\subsection{Examples from \classifier}
\label{sec:examples_of_char}

Below, we provide example stories for categories of character used in \classifier:\\

\noindent\textbf{\textit{Stylization:}}
“You… heroes…” Synthia hissed, blue flames rising around her, spiraling up her frame, “you have it so easy… always loved, always admired… given the powers of a god…”

Confused and cringing, Mara replied, “That’s not true.”

As the witch exploded her magic at Mara, she screeched, “Yes it is!” Quickly, Mara flew upward, using her axe as a shield against the flames, and dodging oncoming winds. Though, no matter how much she moved, Synthia refused to give up, as she detested Mara’s denial.

Flipping through her spellbook, Lux remarked with an awkward laugh, “That’s so weird to say.”

While combatively strumming his guitar, Robbie stopped his humming to agree, “Right? She doesn’t know our lives, we were just sent to fight her.” When he got distracted, his smog slightly shrunk. A spirit’s silhouette rose from it, then slapped his shoulder, making him flinch and complain, “Alright, sorry, damn.” Refocusing, Robbie plucked faster, and the previously-annoyed spirit went back to dancing and clacking their castanets.

After ejecting a boiling ball of plasma at Synthia, Onycia called, “Maybe you think we have it easy because you hate everybody. And, with that, everybody hates you.”

“I hate everybody because everybody has WRONGED ME!” Wind waved over Synthia, knocking Onycia back against a tree. With a snarl, Synthia sent flames her way, but Carlo ran over with his shield. Kneeling in front of Onycia, he casted lightning bolts from his fingertips, which jetted around until they struck Synthia, making her scream and fall to her knees. Though, she recovered surprisingly quickly, rising back to her feet.

Sliding up to Onycia, Tjinfalk muttered, “People have wronged me, but I don’t hate everyone.”

She replied, “I know, love, she’s just strange.”

While Carlo stayed in front of them, Tjinfalk aimed his heat gun at Synthia, balancing the barrel over the shield. Since she was close by, he kept the Distance stat low, but he raised Heat and TIme to max, as she’s proven to be hard to hit, and even harder to knock out. 

Nearby, Synthia was shooting flaming daggers at Lucina, but she was effortlessly and elegantly striking them away with her sword, while saying, “You really think you have it harder than everybody here?”

Sending multiple daggers at once, Synthia growled, “I KNOW I do!”

Lucina grabbed a tree branch above her, pulling herself up to avoid the attack, then jumped back down to nick Synthia across the shoulder. As Synthia gripped the slice, Lucina explained, “Well, Robbie’s dead, I’m homeless, and Tjin hasn’t been to his home planet in six years. We’re not all miserable bums, but that simply scratches the surface. We do not have it easy.”

Catching her breath, the witch scowled, questioning, “You’re homele-” before she could finish, a wide beam of white light obliterated her head, then slowly traced down her body, burning her to a crisp. After about a minute, the beam stopped, and Synthia was nothing more than a pile of ash. 

Carlo settled his shield and stood up, followed by Tjinfalk and Onycia. 

“I still think you need a better name for that weapon,” Robbie told Tjinfalk, “and I still think it should be Hell’s Sun.”

Mara chimed in, “And I still agree!”\\

\noindent\textbf{\textit{Wholeness:}}
It took a lot of convincing to get me to open the door. A lot of cajoling, promises of safety, and patient words of wisdom. Of course, none of that would break me- no, that would take a far more powerful source. 

My cellphone began to buzz. My mother was calling. The idea of NOT picking up briefly crossed my mind,but my fingers were wiser than my brain. The accept button was hit and the phone was pressed to my ear before I could even heave a full sigh. 

“Mom?…Let them in?!How do you kno- you told them it was okay?! Why didn’t you tell me??…I wouldn’t have run-probably….but why are they here?…yeah I would fucking hope I am not in trouble…….I- yes… but…I…ok…I am sorry for cursing. I will use better vocabulary moving forward but thi-… yes…yes..fine…I hope you are right. I trust you. Okay…. I love you too.”

I put the phone back into my pocket and after a long moment and a few deep breaths, I opened the door- body tensed and expecting the worse. 

And to be sure. The next few hours were no picnic. But the conversation didn’t end up going the way I thought. How could I have guessed demon hunters wanted to recruit a half-demon to their ranks?

I clutched the card the lead hunter had given me, still in a daze. A job offer? With full benefits? It was a fantasy even I, a half demon, couldn’t comprehend. 

I was still in my shocked state when my mom came home, clutching bags from my favorite restaurant. She grinned when she saw me.

“I see you had quite the interesting conversation with my old friends when I was gone. Come help me set the table. I got you pie.”

I was on my feet and grabbing plates before she even asked, but froze when I heard her words. 

“Friends?” Is that had she known? Did she send them? Did she tell them about me? Why was she friends with demon hunters when she had a half demon child? Did she know them before or after meeting dad? Did-

Before my mind could race more my mom cleared her throat in a conspicuous manner and nodded her head to the plates in my hand. I jumped and scrambled to my task and she began pulling food out of the bag. 

“Friends yes- Jonah and Bess are two are the best people I knew back in the day, though I guess you could called them “Old colleagues.” instead…” 

She set the pie in the center of the table and sat down. I was about to sit as well when her last words registered and my body opted to collapse into the chair instead. 

“Old…colleagues? But they- you- I… wait.”

My mom, bless her heart, began to chuckle. “Come on darling and eat your dinner. I suppose it’s time I told you the truth about how I met your father….”\\

\noindent\textbf{\textit{Literalness:}}
The year was 2075 when the first known case of reverse time travel was detected—a historical figure illegally entering the present to rectify perceived misconceptions about their lives. The culprit? None other than Socrates, the Greek philosopher whose legacy of wisdom seemed irrevocably tied to his penchant for annoyance.

The sun was barely peeking over the horizon as Lydia, an entry-level archivist at the Temporal Regulatory Agency, sipped her coffee and scanned through yesterday's reports. A flicker of light caught her attention on the screen: a spike in unauthorized temporal energy in Athens, Greece. She leaned in closer, frowning. 

“Not again,” she muttered. Just last week, they’d dealt with Cleopatra insisting on an audience at the Metropolitan Museum of History to discuss inaccuracies in her portrayal. And now Socrates? Of all the historical figures to have spiraled through the use of illegal time travel technology, why did it have to be him?

“Hey! Lydia!” Her colleague Brad’s voice echoed through the office, pulling her from her absorption with the report.

“What?” she replied, not bothering to look at him.

“You wouldn’t believe what just landed in our inbox.” He approached, pointing at the screen.

She raised an eyebrow and turned to face him. “Is it more deranged requests for interviews from ancient leaders? Because I’m done with that. Last time, I almost brought Joan of Arc an answer key for her exam on medieval warfare.”

Brad chuckled, leaning over to make sure the screen displayed the right document. “It’s actually worse. Socrates just sent us a manifesto. He wants to debate the philosophy of ‘falsifying history’ with historians, and the worst part? He’s demanding a public forum.”

Lydia sighed, rubbing her temples. “Great. Just what we need—a debate with a dead philosopher who thinks he’s a living Wikipedia. I mean, how can he even argue? He doesn’t have the benefit of knowing how much has changed since his time.”

But Brad was already animated, his excitement bubbling over. “Think about it! Socrates! The Socratic Method applied to modern problems! It could lead to some killer insights.”

“Or it could lead to hours of semantic drudgery with him trying to philosophically dissect why we called him ‘annoying’ in the twenty-fourth century.” 

Nevertheless, the very next day, Lydia found herself at the public forum. The hall was filled with history enthusiasts and some rather bemused scholars. Lydia couldn’t shake the feeling that they had opened Pandora’s box—or in his case, gave him access to a digital tablet.

When Socrates finally appeared, he looked remarkably unchanged from the statues that adorned classical history texts—bald and bearded, with twinkling eyes that seemed to question everything. The man was completely unfazed by the centuries of progress that had passed since he ingested Hemlock. Instead, his time-traveling escapade had granted him the fervor of a social media influencer and the gravitas of a philosopher, all mashed together.

“Greetings, citizens!” he proclaimed, arms wide. “Let us engage in dialogue about the calamity that is your perception of my trials!”

Lydia rolled her eyes. There was already a debate brewing in the back of the room, and with every question taken, he spiraled deeper into intellectual gymnastics.

“Your records proclaim I was coerced into drinking poison, yet that is but a singular interpretation!” Socrates exclaimed dramatically, raising a finger. “I suggest that, instead, I was willing to embrace true knowledge rather than live a life without virtue.”

The audience was now split; half riled up, eagerly adding to the discussion, while the other half seemed to be channeling their inner “please just leave” attitude through furtive glances at the exits. Lydia had reached her limit.

“Okay, wait! Let’s call a time-out!” she interrupted, forcing her voice to be heard amidst the rising cacophony of pleasure and discontent. “Socrates, what’s the purpose of this? Is it to ensure your historical image is polished, or are you genuinely seeking to educate?”

He turned his deep-set eyes onto her, seeming genuinely puzzled. “Is there a difference?”

“There’s a massive difference!” She exhaled sharply, staring him down. “You’re here now; you’ve seen how people live and learn. Isn’t it about understanding that we can learn from history without erasing it?”

Silence enveloped the room, where seconds stretched into eternity before Socrates nodded slowly, visibly contemplating her words. “Your reasoning holds merit, dear Lydia. Perhaps I’ve aimed too high—and here is the conclusion: to insist on an iron grip of the past may obstruct the flow of wisdom in the present.”

With that, the philosopher accepted a genuine dialogue rather than a correction, and the rest of the evening unfolded into a surprisingly open and generous exchange. Lydia realized that beneath Socrates' seemingly obnoxious airs lay a passion for exploration. 

By the time the sun dipped low, the forum had transformed. It was no longer a battle of ideologies but a sharing of thoughts—a retroactive harmony that was delightful in unexpected ways. 

Lydia leaned back, a smile creeping onto her face, acknowledging that while Socrates could be a jerk, sometimes all it took was a little patience and the willingness to engage honestly. She could only hope that the next reverse traveler would be as accommodating.\\

\noindent\textbf{\textit{Complexity:}}
In the dimly lit chambers of the Infernal Court, echoes of souls in torment reverberated through the stone walls. Just beyond the door, a sign read “Court of Appeals,” scrawled in jagged letters dripping crimson ink, exuding an aura of both dread and hope. Souls were gathered in a loose formation, whispering among themselves, their faces gaunt and hollow, figurative shadows of the once vibrant beings they had been.

At the center of the room sat a massive obsidian table where three judges, cloaked in robes as black as the void, presided. Their faces were obscured by hoods, but their glowing eyes pierced through, exuding an otherworldly judgment that sent shivers of anticipation down the spines of the petitioners. 

“You may approach,” intoned the chief judge, his voice like rustling leaves in a sultry wind. The first soul stepped forward, trembling. The whispers subsided as he began to plead his case. He was met with indifference, and almost snickered at, as he described his life filled with petty crimes and trivial grievances. In no time, the gavel had slammed—denial.

Next came a woman, rage burning in her heart. She spoke of injustice and betrayal, of a life spent in service to others only to have her heart ripped apart by a lover’s treachery. The judges leaned back, their faces cloaked in shadows, and without emotion, they rendered their verdict—denial.

As the hours wore on, the spirits came and went like moths to a flame, each fate met with the same cold dismissal, the cycle of despair indelibly woven into the fabric of the court. But hope remained, as every soul knew there was a chance, however ridiculous, to be among the less than one percent.

Then, with a resounding thud, the heavy door creaked open again, and a soul staggered in, clutching a tattered book to its chest. The specter was different from the rest—torn, yet resolute. Her features bore the uncanny familiarity of a once-renowned author whose stories had touched millions, now rendered unrecognizable.

Alessia, they whispered. Would this be the tale that turned the tide? Would her words, now eager to be freed, touch the judges’ hardened hearts?

“Your Honor,” she began, voice quaking, “I come not to plead for myself, but for the stories I left behind. Books that lie unfinished, characters whose destinies were stolen from them as much as mine was. I plead for their completion, so that my sins—undoing lives through ignorance—might serve a greater purpose.”

The judges shifted slightly, intrigued despite themselves. There was something about her presence, a spark of life that seemed relentless in the face of despair. The chief judge leaned forward, his eyes sharp as daggers.

“And what makes you worthy of our attention?” His tone dripped with skepticism.

Alessia opened her book, pages fluttering in the spectral breeze. “In this volume are the musings of lost souls—a collection titled ‘The Orchard of Regrets.’ Each tale, a life misplaced or misused, each regret a fruit never harvested. Allow me the time to finish this work, and upon its completion, I will accept whatever fate awaits me.”

The judges exchanged glances, a flickering light of curiosity breaking the shadow of their eternal gloom. They were no strangers to ambition yet had grown weary of hollow aspirations. What Alessia proposed was unusual; souls were not known for their selflessness.

“Finish this…book,” the chief judge mused, almost to himself. “But heed this: if you do not succeed, you relinquish your right to another appeal. This volume must evoke emotion, challenge, transformation. Choose your characters wisely.”

She nodded fervently, determination burning in her chest. “I will not falter.”

Days turned into weeks, and Alessia wrote furiously in that desolate chamber, exposing layers of her soul with every word. The characters began to breathe, regretting their unmade choices, mourning paths not taken or lessons unlearned. Souls found solace in her strokes of pen—their stories illuminated in vivid detail, their sorrows transmuted into heartache and beauty.

And as the final draft began to manifest, a peculiar transformation occurred. The despairing torments dulled, the cries of anguish outside the court softened, replaced by the whispers of hope. The three judges observed as the stories twinkled with the essence of all that had been lost and all that could still be reclaimed.

Finally, Alessia approached the obsidian table with the completed manuscript. “Your Honor,” she declared resolutely, “the words are alive. I’ve laid bare my heart and the hearts of those I wronged. I’ve sown understanding where once there was neglect. I ask once more for the chance to redeem myself.”

Time hung heavy in the air, as if the court itself had drawn breath. The judges leaned closer under the dim light, grasping the weight of a legacy that had been reborn.

“I will permit one soul to be released,” the chief judge uttered at last, “but only at your command.”

With a heart fuller than it had ever been, Alessia opened her arms wide. “Let it be for those who wish to reclaim their hopes. Let them run free. And may my story become the seed of their new beginnings.”

As she spoke the final words, a brilliant light shattered the darkness, and the chamber was filled with ethereal laughter. The faces of the souls transformed, each one filled with newfound clarity and purpose as they stepped forward, soaring past the gates of despair, into a world that awaited their return.

And from that moment on, the stories of redemption began to ripple through the sea of lost souls—not every journey would end in release, but every tale would now carry the possibility of eternal hope. Each soul remembered, each life honored, and for Alessia, there would always be a space in history, waiting for a word to be written. The Court of Appeals would henceforth whisper her name, and in the depths of hell, a glimmer of humanity lingered on the horizon.

99 percent of the time, failure reigned. But that 1 percent, that shining anomaly? It transformed the world.\\

\noindent\textbf{\textit{Transparency:}}
The year was 2075 when the first known case of reverse time travel was detected—a historical figure illegally entering the present to rectify perceived misconceptions about their lives. The culprit? None other than Socrates, the Greek philosopher whose legacy of wisdom seemed irrevocably tied to his penchant for annoyance.

The sun was barely peeking over the horizon as Lydia, an entry-level archivist at the Temporal Regulatory Agency, sipped her coffee and scanned through yesterday's reports. A flicker of light caught her attention on the screen: a spike in unauthorized temporal energy in Athens, Greece. She leaned in closer, frowning. 

“Not again,” she muttered. Just last week, they’d dealt with Cleopatra insisting on an audience at the Metropolitan Museum of History to discuss inaccuracies in her portrayal. And now Socrates? Of all the historical figures to have spiraled through the use of illegal time travel technology, why did it have to be him?

“Hey! Lydia!” Her colleague Brad’s voice echoed through the office, pulling her from her absorption with the report.

“What?” she replied, not bothering to look at him.

“You wouldn’t believe what just landed in our inbox.” He approached, pointing at the screen.

She raised an eyebrow and turned to face him. “Is it more deranged requests for interviews from ancient leaders? Because I’m done with that. Last time, I almost brought Joan of Arc an answer key for her exam on medieval warfare.”

Brad chuckled, leaning over to make sure the screen displayed the right document. “It’s actually worse. Socrates just sent us a manifesto. He wants to debate the philosophy of ‘falsifying history’ with historians, and the worst part? He’s demanding a public forum.”

Lydia sighed, rubbing her temples. “Great. Just what we need—a debate with a dead philosopher who thinks he’s a living Wikipedia. I mean, how can he even argue? He doesn’t have the benefit of knowing how much has changed since his time.”

But Brad was already animated, his excitement bubbling over. “Think about it! Socrates! The Socratic Method applied to modern problems! It could lead to some killer insights.”

“Or it could lead to hours of semantic drudgery with him trying to philosophically dissect why we called him ‘annoying’ in the twenty-fourth century.” 

Nevertheless, the very next day, Lydia found herself at the public forum. The hall was filled with history enthusiasts and some rather bemused scholars. Lydia couldn’t shake the feeling that they had opened Pandora’s box—or in his case, gave him access to a digital tablet.

When Socrates finally appeared, he looked remarkably unchanged from the statues that adorned classical history texts—bald and bearded, with twinkling eyes that seemed to question everything. The man was completely unfazed by the centuries of progress that had passed since he ingested Hemlock. Instead, his time-traveling escapade had granted him the fervor of a social media influencer and the gravitas of a philosopher, all mashed together.

“Greetings, citizens!” he proclaimed, arms wide. “Let us engage in dialogue about the calamity that is your perception of my trials!”

Lydia rolled her eyes. There was already a debate brewing in the back of the room, and with every question taken, he spiraled deeper into intellectual gymnastics.

“Your records proclaim I was coerced into drinking poison, yet that is but a singular interpretation!” Socrates exclaimed dramatically, raising a finger. “I suggest that, instead, I was willing to embrace true knowledge rather than live a life without virtue.”

The audience was now split; half riled up, eagerly adding to the discussion, while the other half seemed to be channeling their inner “please just leave” attitude through furtive glances at the exits. Lydia had reached her limit.

“Okay, wait! Let’s call a time-out!” she interrupted, forcing her voice to be heard amidst the rising cacophony of pleasure and discontent. “Socrates, what’s the purpose of this? Is it to ensure your historical image is polished, or are you genuinely seeking to educate?”

He turned his deep-set eyes onto her, seeming genuinely puzzled. “Is there a difference?”

“There’s a massive difference!” She exhaled sharply, staring him down. “You’re here now; you’ve seen how people live and learn. Isn’t it about understanding that we can learn from history without erasing it?”

Silence enveloped the room, where seconds stretched into eternity before Socrates nodded slowly, visibly contemplating her words. “Your reasoning holds merit, dear Lydia. Perhaps I’ve aimed too high—and here is the conclusion: to insist on an iron grip of the past may obstruct the flow of wisdom in the present.”

With that, the philosopher accepted a genuine dialogue rather than a correction, and the rest of the evening unfolded into a surprisingly open and generous exchange. Lydia realized that beneath Socrates' seemingly obnoxious airs lay a passion for exploration. 

By the time the sun dipped low, the forum had transformed. It was no longer a battle of ideologies but a sharing of thoughts—a retroactive harmony that was delightful in unexpected ways. 

Lydia leaned back, a smile creeping onto her face, acknowledging that while Socrates could be a jerk, sometimes all it took was a little patience and the willingness to engage honestly. She could only hope that the next reverse traveler would be as accommodating.\\

\noindent\textbf{\textit{Dynamism:}}
"Until Help arrives, stay hidden or barricaded, do not engage any of the subjects at any cost. They are hostile and highly infectious. We repeat, DO NOT ENGAGE."

The alarm bleared through many electronic signals. It started so soon that no one had true information about this situation. We only knew fear.

"Until Help arrives, stay hidden or barricaded, do not engage any of the subjects at any cost. They are hostile and highly infectious. We repeat, DO NOT ENGAGE."

The alarm repeats, the commentator is tired and it shows... Great distress etched in their face, as if a great burden had been forced upon them. Some forbidden knowledge, or worse...

"Until Help arrives, stay hidden or barricaded, do not engage any of the subjects at any cost. They are hostile and highly infectious. We repeat, DO NOT ENGAGE."

Three times it had repeated... Nothing happened, yet...

We grow weary of them. We resent the authorities forcing us into the confinement of our homes.

"Until Help arrives, stay hidden or barricaded, do not engage any of the subjects at any cost. They are hostile and highly infectious. We repeat, DO NOT ENGAGE."

It has been a day since the start of the signal. I heard shots and screams... Being alone is bad during these trying times. It is worse hearing the mad ramblings of some people. Omens of terrible times coming, a test from the heavens... The return of the old gods. Being confined had eroded their cognitive abilities...

We can't keep this.

"Until Help arrives, stay hidden or barricaded, do not engage any of the subjects at any cost. They are hostile and highly infectious. We repeat, DO NOT ENGAGE."

And now, I hear someone or something hit my barricaded windows. It started strong, but now it has slowed down... It was almost rhythmic... Soothing.

Until it wasn't

"Until Help arrives, stay hidden or barricaded, do not engage any of the subjects at any cost. They are hostile and-"

An injured woman managed to break into my home.

She is bleeding from the gunshots... Her skin presents some growths, like moss or lichen. Science was never my strong school course...

But there is something that attracts me. A smell, or maybe it is the basic human empathy.

Since she broke into my home, I turned off my radio... I grew bored of this doom-saying. And I see her again. Something compels me to hold her... And protect her.

To try and heal her injuries.

This was my doom.\\

\noindent\textbf{\textit{Closure:}}
The world had crumbled under the weight of the undead. Cities that once bustled with life were now husks, echoing only with the shuffling sounds of the infected. When the zombie apocalypse descended like a dark cloud, most people rushed to gun stores, believing firepower would be their salvation. But not Lucas. When the chaos erupted, he found himself drawn to a time long past, a place where metal was forged into legends and honor marred by few scratches.

In a sleepy town at the edge of civilization, Lucas had discovered an old medieval armory tucked away behind a forgotten alley. Its wooden door whispered secrets of battles long settled, and upon breaking in, he breathed the air saturated with history. There, the armory stood proud and untouched: knights' helmets, greaves, and swords elegantly arranged like art begging for a master. Lucas knew in that moment what he had to do.

He donned a full suit of plate armor, the metal cool against his skin as he cinched the leather straps tight. It was cumbersome yet invincible, every piece a testament to the craftsmanship of an era defined by valor. He felt a surge of power—more than just the weight of the armor, but the indomitable spirit of the knight who once wore it.

Clutching a long sword with a cruciform hilt, Lucas set out onto the streets of his neighborhood. The clamor of the armor accompanied him, a reminder that he was a bulwark against the horrors lurking beyond his door. The streets were desolate, houses boarded, and windows shattered, yet he was determined to reclaim his home from the marauding dead.

The first group of zombies he encountered lurked in the shadows of an abandoned car lot—four of them, hands grasping desperately at the ground, heads lolling. Adrenaline surged through him as he unleashed a harsh battle cry. The crash of metal against pavement startled the creatures, their lifeless eyes snapping to attention.

With precision born of fabled conflicts long ago, Lucas swung his sword. The blade sang through the air, catching the first zombie square in the neck, cleaving through the decayed flesh with just one stroke. The body crumpled to the ground, lifeless once more. The smell of rot mingled with the sweet scent of bravery; he felt alive.

The remaining zombies roared, stumbling forward in a mindless frenzy. Lucas maneuvered effortlessly, the weight of his armor grounding him, preventing the undead from tumbling him over. He sidestepped the nearest attacker, his sword wheeling like a tempest, catching another zombie in the ribs. The momentum of his swing sent another stumbling backward, and quenching the instinct to retreat, he pursued with zeal.

One after another, the creatures fell before him. Reflected sunlight glinted off his polished armor while the thrill of battle coursed through his veins. He felt as though he had been transported to a different age, where his every strike rang out like a proclamation—he was not to be trifled with.

As he continued through the streets, clearing his neighborhood, Lucas discovered remnants of life amidst the desolation. An overturned bicycle, a child’s drawing, a dog collar—small reminders of the innocence that had been cast aside. And in these elements, his determination grew. He was not just fighting zombies; he was reclaiming the echoes of laughter and the fleeting moments of joy that used to fill the sidewalks.

Hours later, with the sun setting over the horizon and the air thick with the scent of spent decay, Lucas stood upon what used to be his front lawn. It was littered with bodies, but none rose again. He had cleared the immediate vicinity, a heroic effort that left him breathless and triumphant.

He unsheathed his sword, allowing it to rest against his shoulder as he surveyed the familiar surroundings. Such tranquility now felt surreal; he was not just a fool in armor; he was a guardian, a protector embodying the spirit of every hero that had ever roamed this world with honor.

As he made his way back into the house, Lucas now carried more than just firepower; he possessed hope—a weapon forged in steel. And as long as he stood between the shadows and the light, he would fight to ensure that this was not merely an end, but a beginning waiting to be written anew.

\section{Evaluating Prompting Methods}
\subsection{Templates}
\label{append:evaluating_prompting_methods}

In order to find best templates for \classifier, we try out numerous formats. We try several variations of zero-shot and ICL learning, where category-pairs are classified one-pair-at-time, all at once in a single prompt, and on a Likert scale (values 0-2 are bucketed into category A, values 3-5 are bucketed into category B). In addition to using definitions for category-pairs (see Table \ref{tab:category_pairs_defs}), we try replacing the definitions with descriptive synonyms (akin to a thesaurus lookup) as well as concatenating the definitions with the synonyms. We determine that the templates using only the definitions work best.

Below we provide the best prompt templates from our experiments for classifying categories for characters. 
In \classifier\ we use \textbf{Zero-shot} for \textit{dynamism} and \textit{closure}, 
\textbf{ICL-basic} for \textit{literalness}, \textbf{ICL-interleaved} for \textit{coherence}, \textit{wholeness}, \textit{complexity}, and \textit{transparency}, and
\textbf{ICL-repeated} for \textit{stylization}.

Note: for all ICL templates, shots are chosen randomly from the gold labels/explanations for each inference call. In this way, we avoid unduly biasing the classification towards a single story in each category.

\begin{mybox}[Zero-shot]
You are an expert literary analyst. Given a short narrative, honestly evaluate it and determine the character type of the protagonist.  \\

We define the following character types:\\
\texttt{\{a\_def\}} 
\texttt{\{b\_def\}}\\

Now, given the following narrative, classify the protagonist. Respond in valid JSON only, with two fields:  
\{"explanation": "a short justification under 50 tokens", "solution": "A or B"\}  \\

Narrative: \texttt{\{story\}}
\end{mybox}

\begin{mybox}[ICL-Basic]
You are an expert literary analyst. Given a short narrative, honestly evaluate it and determine the character type of the protagonist.  \\

We define the following character types:  \\
\texttt{\{a\_def\}}  
\texttt{\{b\_def\}}  \\

Example 1: \texttt{\{a\_story\}}  \\
Solution 1: \{``explanation'': \texttt{\{a\_explain\}}, ``solution'': ``A''\}\\

Example 2: \texttt{\{b\_story\}}  \\
Solution 2: \{``explanation'': \texttt{\{b\_explain\}}, ``solution'': ``B''\}\\ 

Now, given the following narrative, classify the protagonist. Respond in valid JSON only, with two fields:  
\{``explanation'': ``a short justification under 50 tokens'', ``solution'': ``A or B''\} \\

Narrative: \texttt{\{story\}}
\end{mybox}

\begin{mybox}[ICL-Interleaved]
You are an expert literary analyst. Given a short narrative, honestly evaluate it and determine the character type of the protagonist. We define the following character types: \\

\texttt{\{a\_def\}}  \\
Example: \texttt{\{a\_story\}}\\  
Solution: \{``explanation'': \texttt{\{a\_explain\}}, ``solution'': ``A''\}\\  

\texttt{\{b\_def\}}\\  
Example: \texttt{\{b\_story\}}\\  
Solution: \{``explanation'': \texttt{\{b\_explain\}}, ``solution'': ``B''\}\\  

Now, given the following narrative, classify the protagonist. Respond in valid JSON only, with two fields:  
\{``explanation'': ``a short justification under 50 tokens'', ``solution'': ``A or B''\}  \\

Narrative: \texttt{\{story\}}
\end{mybox}

\begin{mybox}[ICL-Repeated]
You are an expert literary analyst. Given a short narrative, honestly evaluate it and determine the character type of the protagonist. \\ 

We define the following character types:\\  
\texttt{\{a\_def\}}  
\texttt{\{b\_def\}}  \\

Example 1: \texttt{\{a\_story\}}\\  
Solution 1: \{"explanation": "\texttt{\{a\_explain\}}", "solution": "A"\}\\  

Example 2: \texttt{\{b\_story\}}\\  
Solution 2: \{"explanation": "\texttt{\{b\_explain\}}", "solution": "B"\}\\  

Reminder: The character types are:  
\texttt{\{a\_def\}}  
\texttt{\{b\_def\}} \\ 

Now, given the following narrative, classify the protagonist. Respond in valid JSON only, with two fields:  
\{"explanation": "a short justification under 50 tokens", "solution": "A or B"\}  \\

Narrative: \texttt{\{story\}}
\end{mybox}

\begin{mybox}[Combined]
You are an expert literary analyst. Given a short narrative, evaluate the protagonist across eight dimensions of characterization.  \\

Here are the definitions:  \\

1. Stylization vs. Naturalism  \\
\texttt{\{a\_def\}}
\texttt{\{b\_def\}}  \\

2. Coherence vs. Incoherence  \\
\texttt{\{a\_def\}}
\texttt{\{b\_def\}}  \\

3. Wholeness vs. Fragmentariness \\ 
\texttt{\{a\_def\}}
\texttt{\{b\_def\}}  \\

4. Literalness vs. Symbolism \\ 
\texttt{\{a\_def\}}
\texttt{\{b\_def\}}  \\

5. Complexity vs. Simplicity  \\
\texttt{\{a\_def\}}
\texttt{\{b\_def\}}  \\

6. Transparency vs. Opacity  \\
\texttt{\{a\_def\}}
\texttt{\{b\_def\}}  \\

7. Dynamism vs. Staticism  \\
\texttt{\{a\_def\}}
\texttt{\{b\_def\}}  \\

8. Closure vs. Openness \\ 
\texttt{\{a\_def\}}
\texttt{\{b\_def\}}  \\

Classify the protagonist in the following narrative.  
Respond ONLY in valid JSON with exactly this format. Each category must be either "A" or "B" (not words):\\  

\{``category1'': ``A or B'', ``category2'': ``A or B'', ``category3'': ``A or B'', ``category4'': ``A or B'', ``category5'': ``A or B'', ``category6'': ``A or B'', ``category7'': ``A or B'', ``category8'': ``A or B''\}  

Narrative: \{story\}
\end{mybox}

\begin{mybox}[Likert-Scale]
You are an expert literary analyst. Given a short narrative, honestly evaluate it and determine the character type of the protagonist.  \\

We define the following character types:  \\
\texttt{\{a\_def\} } 
\texttt{\{b\_def\} } \\

Given the following narrative, rate the protagonist on a scale from 0 to 5, where:  \\
- 0 means the protagonist clearly matches category A. \\ 
- 5 means the protagonist clearly matches category B.  \\
- Numbers between 1–4 represent varying degrees between A and B.  \\

DO NOT GIVE ANY EXPLANATION. Respond with ONLY the number (0, 1, 2, 3, 4, or 5).  \\

Narrative: \texttt{\{story\}}
\end{mybox}

\subsection{Test Set Inter-Rater Agreement}
\label{sec:inter-rater-agreement}

The \classifier\ test set is manually annotated by 4 annotators, 2-3 annotators per story. All annotators are native English speakers and current undergraduate/graduate students in the USA. To train them for the task, they are provided descriptions and definitions of the categories, including those used in our prompt templates and the source paper \cite{hochman1985character} where they are originally described. Each story is initially annotated by 2 in-person reviewers. For any disagreement a third annotator chooses the final label.
Table~\ref{tab:annotator-agreement} provides the breakdown of Cohen's kappa inter-rater agreement and F1-macro scores. 

Note: Human annotations are very difficult, time-consuming and expensive. This is because annotators need to be trained to be experts in the relevant part of literary theory and read between the lines to understand character portrayal. For 100 stories, we have a total of 800 gold annotations. While this test set size is small, we attempt to ensure that it is high quality.

\begin{table}[t]
\centering
\footnotesize
\begin{tabular}{lcc}
\toprule
\textbf{Category} & \textbf{$\kappa$} & \textbf{F1-macro} \\
\toprule
\textit{Stylization}   & 63.7 & 87.66\\
\textit{Coherence}     & 46.1 & 69.23\\
\textit{Wholeness}     & 52.0 & 78.44\\
\textit{Literalness}   & 55.2 & 89.39\\
\textit{Complexity}    & 60.9 & 83.67\\
\textit{Transparency}  & 63.2 & 88.84\\
\textit{Dynamism}      & 83.6 & 87.1\\
\textit{Closure}       & 56.0 & 92.79\\
\bottomrule
\end{tabular}
\caption{Inter-rater agreement, shown with Cohen's kappa and F1-macro score as percentages.}
\label{tab:annotator-agreement}
\end{table}

\begin{table*}[ht]
\centering
\footnotesize
\begin{tabular}{lcccc|cc}
\toprule
\textbf{Category} & \textbf{Zero-shot} & \textbf{ICL ``basic''} & \textbf{ICL ``dispersed''} & \textbf{ICL ``repeated''} & \textbf{Likert} & \textbf{Single-prompt} \\
\toprule
\textit{Stylization}   & 43.18 & 66.02 & 67.84 & \textbf{68.47} & 64.40 & 44.55 \\
\textit{Coherence}     & 50.76 & \textbf{62.99} & 59.28 & 59.28 & 48.72 & 48.98 \\
\textit{Wholeness}     & 52.81 & \textbf{57.48} & 56.71 & 57.48 & 61.02 & 54.83 \\
\textit{Literalness}   & 45.28 & 49.27 & \textbf{60.81} & 56.00 & 61.13 & 65.23 \\
\textit{Complexity}    & 46.52 & \textbf{61.69} & 61.69 & 54.78 & 46.52 & 53.15 \\
\textit{Transparency}  & 39.15 & \textbf{60.10} & 55.56 & 45.65 & 45.95 & 57.48 \\
\textit{Dynamism}      & \textbf{84.37} & 76.62 & 76.19 & 79.32 & 56.10 & 77.63 \\
\textit{Closure}       & \textbf{75.12} & 65.99 & 70.86 & 72.78 & 67.99 & 57.93 \\
\bottomrule
\end{tabular}
\caption{Comparison of templates for each category (F1-macro scores). Templates used in \classifier\ are in bold. Results are statistically significant using Bootstrap Test \cite{dror2018hitchhiker}. While comparing three runs, the largest standard deviation is 2.12\% (for closure).}
\label{tab:experiment-f1}
\end{table*}

\begin{table}[h]
\centering
\footnotesize
\begin{tabular}{lcc}
\toprule
\textbf{Category} &\textbf{Llama-70B} & \textbf{GPT4o-mini} \\
\toprule
\textit{Stylization}   & 68.47 & 66.67 \\
\textit{Coherence}   & 62.99 & 61.60 \\
\textit{Wholeness}   & 57.48 & 58.88 \\
\textit{Literalness}   & 60.81 & 63.53 \\
\textit{Complexity}   & 61.69 & 59.35 \\
\textit{Transparency}   & 60.10 & 59.35 \\
\textit{Dynamism}   & 84.37 & 78.18 \\
\textit{Closure}   & 75.12 & 39.27 \\
\bottomrule
\end{tabular}
\caption{Comparison of Llama-70B vs. GPT4o-mini, using \classifier\ templates (F1-macro scores, in percent).}
\label{tab:llama_gpt}
\end{table}

\subsection{Results from Experiments with Templates}
\label{sub:results-experiments-templates}

\begin{table}[h]
\footnotesize
\centering
\begin{tabular}{lcccc}
\toprule
\textbf{Categories} & \textbf{Pred A} & \textbf{Pred B} & \textbf{Gold A} & \textbf{Gold B} \\
\midrule
\textit{Stylization}   & 0.50 & 0.50 & 0.72 & 0.28 \\
\textit{Coherence}     & 0.96 & 0.04 & 0.97 & 0.03 \\
\textit{Wholeness}     & 0.88 & 0.12 & 0.78 & 0.22 \\
\textit{Literalness}   & 0.59 & 0.41 & 0.34 & 0.66 \\
\textit{Complexity}    & 0.84 & 0.16 & 0.87 & 0.13 \\
\textit{Transparency}  & 0.94 & 0.06 & 0.85 & 0.15 \\
\textit{Dynamism}      & 0.68 & 0.32 & 0.73 & 0.27 \\
\textit{Closure}       & 0.25 & 0.75 & 0.41 & 0.59 \\
\bottomrule
\end{tabular}
\caption{Percentage of times the LLM predicts label A and B versus the percentage of occurrences of label A and B occurring as gold labels.}
\label{tab:category_scores}
\end{table}

\begin{table}[h]
\centering
\footnotesize
\begin{tabular}{lcc}
\toprule
\textbf{Categories} & \textbf{Human} & \textbf{LLM} \\
\midrule
\textit{Stylization}  & 0.11 & 0.30 \\
\textit{Coherence}    & 0.03 & 0.05 \\
\textit{Wholeness}    & 0.13 & 0.24 \\
\textit{Literalness}  & 0.10 & 0.39 \\
\textit{Complexity}   & 0.09 & 0.19 \\
\textit{Transparency} & 0.06 & 0.15 \\
\textit{Dynamism}     & 0.09 & 0.13 \\
\textit{Closure}      & 0.07 & 0.22 \\
\bottomrule
\end{tabular}
\caption{Percentage of disagreements between human annotators and LLM predictions for each category.}
\label{tab:disagreements}
\end{table}

\begin{figure*}
    \centering
    \includegraphics[width=\linewidth]{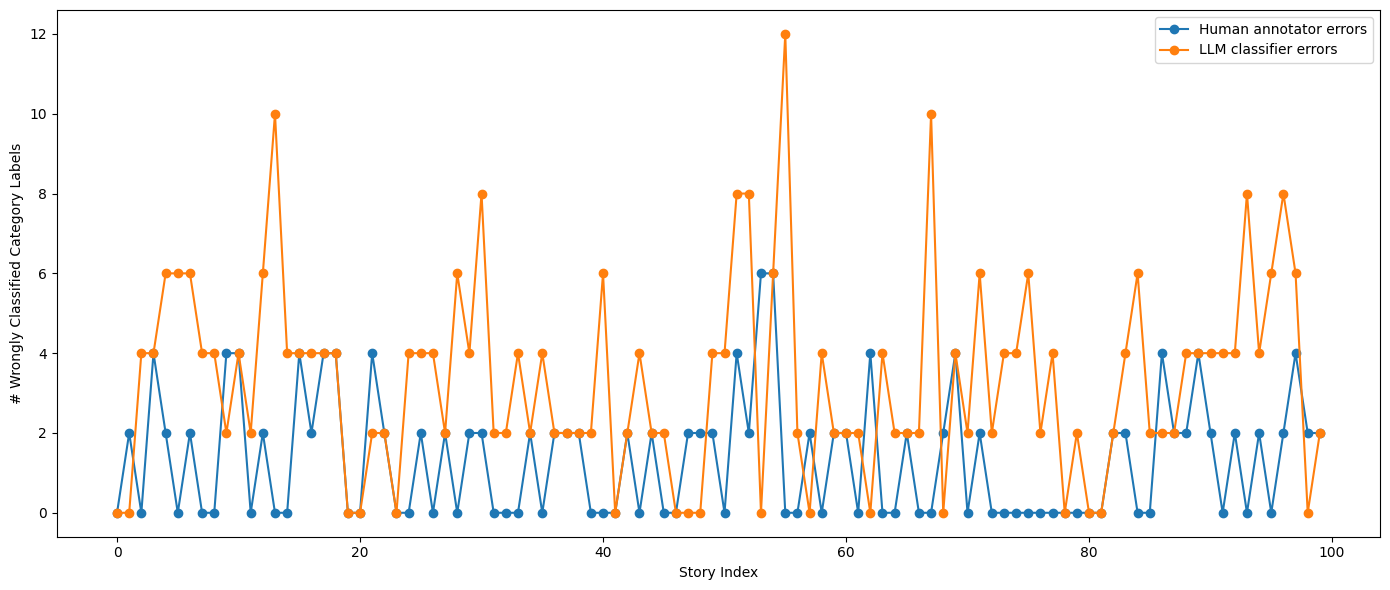}
    \caption{Comparison of wrongly classified category labels for humans (blue) vs. LLMs (orange). The x-axis represents the index of each classified story.}
    \label{fig:tricky-stories}
\end{figure*}

See Table \ref{tab:experiment-f1} for F1-macro scores obtained during experimentation, described in Section \ref{sec:evaluation-framework}, to find best prompt templates for \classifier. See Table \ref{tab:llama_gpt} for a comparison of \texttt{Llama-70B} and \texttt{GPT4o-mini}. We observe that \texttt{Llama-70B} performs comparably to (and sometimes even better than) \texttt{GPT4o-mini}, so we use this model to complete our analysis of LLM vs. human-written characters. 

We perform the following analyses to better understand why the classifier sometimes makes mistakes, even with the best templates:

\begin{enumerate}
    
    \item \textit{Do LLMs prefer certain labels for certain traits?} We count the number of times the LLM predicts a label (A vs. B, where A represents the category, and B represents the opposing category). Table \ref{tab:category_scores} shows the breakdown of the percentages of each label occurrence. LLMs tend to under-classify \textit{stylized} and \textit{closed} characters and over-classify \textit{literal} characters. We determine that these are the trickiest categories to predict.

    \item \textit{Do LLMs and Human annotators share similar disagreements rates?} We count the percentage of disagreements between human annotators and LLM predictions vs. gold labels Table \ref{tab:disagreements}. These columns have a moderately strong/strong positive correlation with Pearson correlation $(r) \approx 0.695$. The results indicate that characters who are difficult or subjective for humans to classify are also difficult for LLMs to classify.

    \item \textit{Do classification errors occur because particular stories/characters are tricky to analyze overall?} Figure \ref{fig:tricky-stories} shows how much a human annotator differs from the final gold label and the error rate of an LLM classifier across all 8 categories for each story. For a few stories, LLMs make errors, while humans make no errors at all. There are also a few cases of humans making classification errors where LLMs classify correctly. Overall, we do not see overarching trends that suggest certain stories are trickier to interpret than others.
\end{enumerate}

\section{Implementation Details}
\label{sec:implementation-details}

\begin{table*}[t]
\centering
\footnotesize
\begin{tabular}{lllll}
\toprule
\textbf{Family} & \textbf{Short Name} & \textbf{Exact Model Name} & \textbf{Size} & \textbf{Knowledge Cutoff} \\
\toprule
GPT & GPT4o-mini & gpt-4o-mini-2024-07-18 & \textit{Unknown} & 09/30/2023\\
\midrule
\multirow{3}{*}{Llama} & Llama-3B & meta-llama/Llama-3.2-3B-Instruct & 3.21B & 12/2023\\
& Llama-8B & meta-llama/Llama-3.1-8B-Instruct & 8B & 12/2023\\
& Llama-70B & meta-llama/Llama-3.3-70B-Instruct & 70B & 12/2023\\
\midrule
\multirow{2}{*}{Phi} & Phi3-4B & microsoft/Phi-3-mini-4k-instruct & 3.8B & 10/2023\\
& Phi3-14B & microsoft/Phi-3-medium-4k-instruct & 14B & 10/2023\\
\midrule
\multirow{2}{*}{Mistral} & Mistral-7B & mistralai/Mistral-7B-Instruct-v0.3 & 7.25B & 02/2023 \\
& Mistral-24B & mistralai/Mistral-Small-24B-Instruct-2501 & 23.6B & \textit{Unknown} \\
\bottomrule
\end{tabular}
\caption{Details about all LLMs used in our experiments.}
\label{tab:model_families}
\end{table*}

Table \ref{tab:model_families} contains the complete list of LLMs with their respective family names, exact model names from HuggingFace library \cite{wolf2019huggingface}, model sizes, and knowledge cutoff dates. Experiments are conducted using OpenAI API for the closed-source model and four 48GB Nvidia RTX A6000 GPUs for all open-source models. Generating 200 stories does not exceed 2 hours per LLM. Inference time for each experiment takes approximately 20 minutes for 200 stories.

We use existing Python packages such as HuggingFace, pandas, re, scikit-learn, and statistics. We scrape the human-written stories from Reddit using the Reddit API Wrapper, PRAW. 

We note that all stories used in \classifier\ are freely available for research purposes according to the Reddit Public Content Policy.

AI is used for minor assistance in coding.

\begin{figure}
    \centering
    \includegraphics[width=\linewidth]{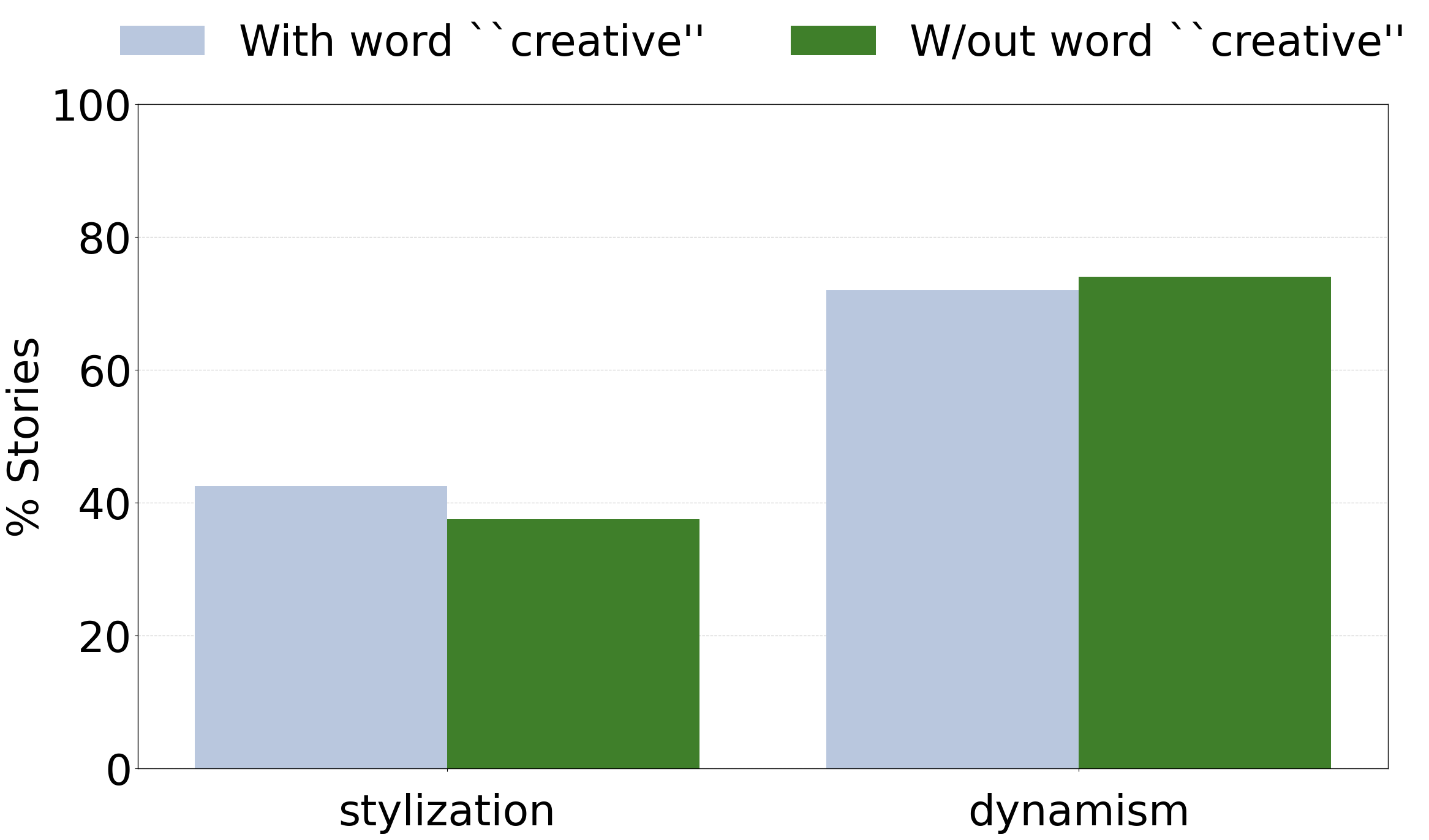}
    \caption{Comparison of types of characters generated when template includes vs. does not include the word ``creative.'' For both \textit{stylization} and \textit{dynamism}, we observe minimal differences. These results indicate that the use of the word ``creative'' does not bias the LLM towards generating significantly more \textit{stylized} and \textit{dynamic} characters.}
    \label{fig:use_of_creative}
\end{figure}

\section{Checking ``Creativity'' in Test Set}
\label{sec:appendix_checking_creativity}

In order to generate the test set, we use a prompt template that includes the phrase ``You are a creative story writer.'' Here, we show that the word ``creative'' does not bias the LLM towards generating more creative (e.g., more dramatic) characters than it would otherwise. From the category-pairs, we focus on \textit{stylization} and \textit{dynamism}. Using a random sampling of 50 writing-prompts, we generate 2 stories per writing-prompt (one where the prompt template includes and one where the prompt template excludes the word ``creative''). We use \texttt{Llama-8B} for story-generation and \texttt{Llama-70B} for \classifier\ classification. As shown in Figure~\ref{fig:use_of_creative}, there is minimal difference in the types of characters generated with respect to \textit{stylization} and \textit{dynamism}. Hence, we conclude that the use of the word ``creative'' does not bias the LLM during story generation.

\section{Comparing Human-Written Stories}
\label{sec:comparining_human_written_stories}

Since the human-written stories from our testset and the human-written stories from the \classifier\ corpus are obtained from different sources (two subreddits), we conduct experiments to ensure the textual domains of the two sets should be indistinguishable. For this experiment, we use in-context learning with \texttt{Llama-70B} to classify each story as belonging to either the testset or to the \classifier\ corpus. We use the following prompt:
\begin{mybox}[Evaluating Domain Comparability]
There are 2 datasets of short stories. Given a new short story, determine which dataset it belongs to.

Dataset 1 example: \{\textit{shots}\}

Dataset 2 examples: \{\textit{shots}\}

New story: \{\textit{story}\}
\end{mybox}

The resulting Macro-F1 score $=24.52\%$. This score is very low, indicating that the model is unable to distinguish stories from the two sources. Thus, we determine that the stories are from the same textual domain.

We also use available software for detecting machine-generated text, including \texttt{Binoculars}\ \cite{hans2024spotting} and Copyleaks AI Detector.\footnote{https://copyleaks.com/ai-content-detector} In this way, we try to choose stories which are primarily written by humans.

\section{Additional Details about \classifier\ Corpus}
\label{sec:addition-details-corpus}

See Table \ref{tab:corpus_statistics} for additional statistics for the \classifier\ corpus.

\begin{table*}[t]
    \centering
    \footnotesize
    \begin{tabular}{llcccc}
        \toprule
        Source Family & Source Name & \# Stories & Avg & Min & Max \\
        \toprule
        \multirow{2}{*}{Human}&\texttt{r/WritingPrompts} & 50 & 661.48$\pm$302.55& 177& 1739\\
        &\texttt{r/shortstories} & 200 & 1502.57$\pm$1117.83 & 494 & 6395 \\
        \midrule
        GPT & GPT4o-mini & 200 & 952.57$\pm$118.13 & 528 & 1503\\
        \midrule
        \multirow{3}{*}{Llama}&Llama-3B& 600 & 686.78$\pm$174.67 & 5 & 1146\\
        &Llama-8B& 600 & 695.69$\pm$122.47 & 17 & 1214\\
        &Llama-70B& 600 & 1027.53$\pm$401.79 & 457 & 1787\\
        \midrule
        \multirow{2}{*}{Phi}&Phi3-4B& 600 & 763.71$\pm$306.16 & 13 & 1561\\
        &Phi3-14B& 600 & 759.29$\pm$175 & 40 & 1445\\
        \midrule
        \multirow{2}{*}{Mistral}&Mistral-7B& 600 & 560.83$\pm$152.44 & 241 & 1462\\
        &Mistral-24B& 600 & 845.06$\pm$267.79 & 358 & 1687\\
        \bottomrule
    \end{tabular}
    \caption{Corpus statistics, including average, minimum, and maximum number of tokens in each story.}
    \label{tab:corpus_statistics}
\end{table*}

\section{Mapping of Genre Labels}
\label{append:mapping_of_genre_labels}

Table \ref{tab:genre_mappings} shows the original Subreddit tags used to indicate human-labeled story genres. We merge and map these genres to our set of 4 broad genres, Domestic Fiction, Romance, Science-Fiction, and Suspense/Thriller.

\begin{table}[t]
    \centering
    \footnotesize
    \begin{tabular}{ll}
        \toprule
        \textbf{Genre (from Subreddit)} & \textbf{Merge with} \\
        \midrule
        Science Fiction & Science-Fiction/Fantasy \\
        Fantasy & Science-Fiction/Fantasy \\
        Realistic Fiction & Domestic Fiction \\
        Horror & Suspense/Thriller \\
        Misc Fiction & Drop \\
        Speculative Fiction & Science-Fiction/Fantasy \\
        Humour & Drop \\
        Romance & Romance \\
        Non-Fiction & Drop \\
        Mystery \& Suspense & Suspense/Thriller \\
        Thriller & Suspense/Thriller \\
        Historical Fiction & Drop \\
        Serial Sunday & Drop \\
        Action \& Adventure & Suspense/Thriller \\
        Urban & Domestic Fiction \\
        Micro Monday & Drop \\
        Off Topic & Drop \\
        Meta Post & Drop \\
        Other & Drop \\
        \bottomrule
    \end{tabular}
    \caption{Genre mappings from subreddit tags to our genres.}
    \label{tab:genre_mappings}
\end{table}

\section{Additional Analysis}

\subsection{More Details about RQ5}
\label{sec:genres-for-human-stories}

\begin{figure}
    \centering
    \includegraphics[width=\linewidth]{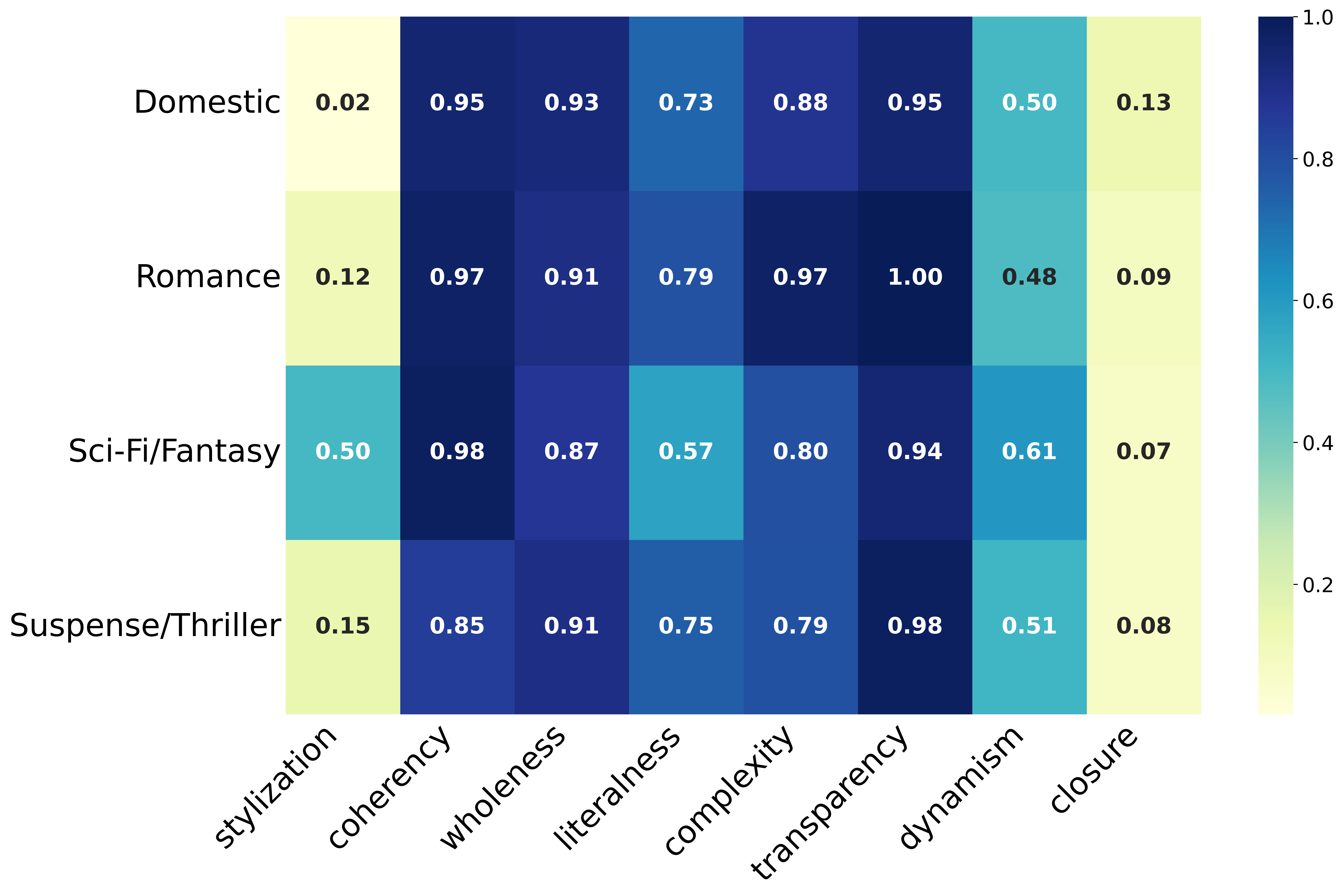}
    \caption{\textbf{RQ5:} Distribution of categories, given each specified genre across human-written stories.}
    \label{fig:rq5-human}
\end{figure}

For Section \ref{sec:analysis} RQ5, in additional to showing the heatmap of the distribution of categories, given each specified genre across all LLM-generated stories, (Figure \ref{fig:rq5}) we provide a corresponding heatmap for all human-written stories in Figure \ref{fig:rq5-human}. We observe that many of the distributions are similar to the distributions of the LLM-generated stories, indicating that when genres are specified, LLM-generated stories are biased toward generating stories with particular categories of characters in alignment with human-written stories. One noticeable difference, however, is that more LLM-generated characters are \textit{closed} compared to human-written characters, further emphasizing our first key takeaway (Section \ref{sec:takeaways}) that LLMs prefer to "play it safe" by ending a story when the character is also completed.

\subsection{More Details about RQ6}
\label{sec:rq6_appendix}
\begin{figure*}[t]
    \centering
    \includegraphics[width=.8\linewidth]{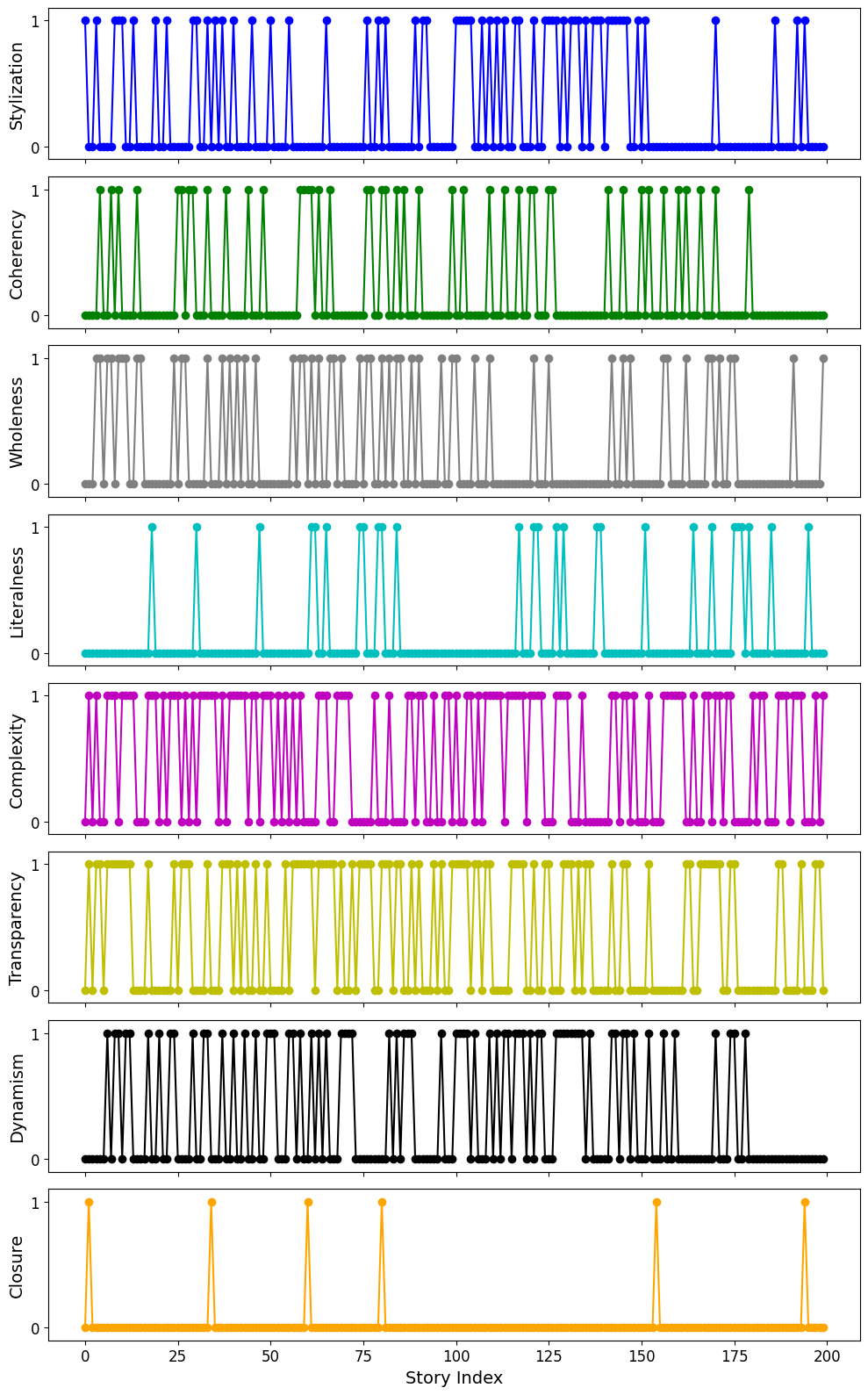}
    \caption{\textbf{RQ6:} Variability of categories across 3 identical inference calls (averaged over all open-sourced LLMs). X-axes are writing-prompt IDs. For a given writing-prompt, y-axes show whether the generated character has identical categories across all 3 inference calls ($y=1$) or if at least one label differs ($y=0$). For \textit{stylization}, \textit{coherency}, \textit{wholeness}, \textit{literalness}, \textit{complexity}, \textit{transparency},
    \textit{dynamism} and \textit{closure}, respectively 30\%, 22\%, 27.5\%, 13.5\%, 53.5\%, 45.5\%, 35.5\% and 3\% of the writing-prompts yield characters with the same label across all inference calls.}
    \label{fig:rq7_all}
\end{figure*}

In Figure \ref{fig:rq7_all} we provide graphs showing variability of categories across 3 separate story generations using the same writing-prompt for all categories.

\end{document}